\documentclass[10pt,twocolumn,letterpaper]{article}

\usepackage[pagenumbers]{cvpr} 

\usepackage[accsupp]{axessibility}
\usepackage{times}
\usepackage{epsfig}
\usepackage{graphicx}
\usepackage{amsmath}
\usepackage{amssymb}
\usepackage{amsthm}

\usepackage{booktabs}
\usepackage{bm}
\usepackage{caption}
\usepackage{makecell}
\usepackage[dvipsnames]{xcolor}
\usepackage[pagebackref,breaklinks,colorlinks]{hyperref}

\usepackage[capitalize]{cleveref}
\crefname{section}{Sec.}{Secs.}
\Crefname{section}{Section}{Sections}
\Crefname{table}{Table}{Tables}
\crefname{table}{Tab.}{Tabs.}

\newcommand{\myparagraph}[1]{\vspace{0.2em}\noindent\textbf{#1}}

\def\cvprPaperID{0000} 
\def\confName{CVPR}
\def\confYear{1920}

\begin{document}

\title{The Wanderings of Odysseus in 3D Scenes}



\newcommand*{\affaddr}[1]{#1} 
\newcommand*{\affmark}[1][*]{\textsuperscript{#1}}
\newcommand*{\email}[1]{\small{\texttt{#1}}}

\author{
Yan Zhang \quad
Siyu Tang\\
\affaddr{ETH Z\"urich}\\
\email{\{yan.zhang,siyu.tang\}@inf.ethz.ch} 
}

\twocolumn[{%
\renewcommand\twocolumn[1][]{#1}%
\maketitle
\begin{center}
    \newcommand{\teaserwidth}{\textwidth}
\vspace{-0.2in}
    \centerline{
    \includegraphics[width=1\linewidth]{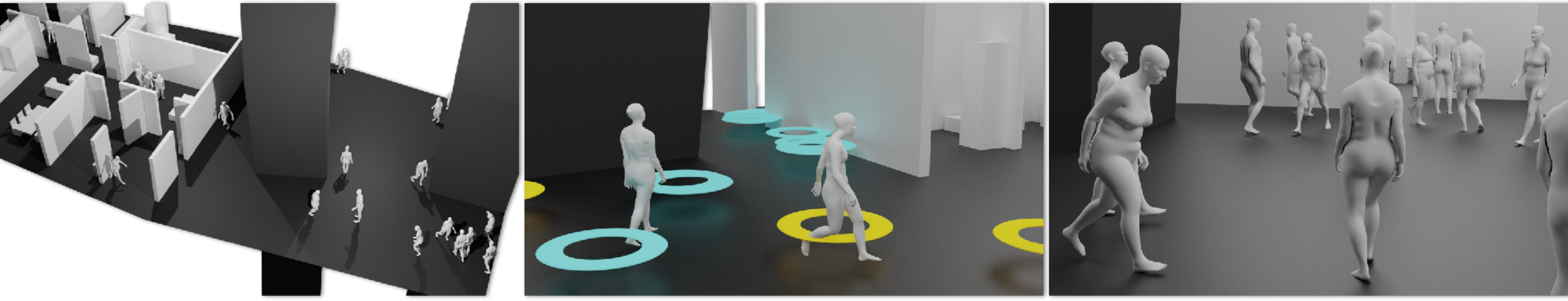}
     }
  \captionof{figure}{We propose {\bf GAMMA}, an automatic and scalable solution, to populate the 3D scene with digital humans, which have 1) varied body shapes, 2) plausible body-ground contact, and 3) realistic and perpetual motions to reach goals (e.g. {\color{BlueGreen} rings} in the middle figure), with the off-the-shelf path-finding algorithm.
  }
\label{fig:teaser}
\end{center}%
}]

\maketitle

\begin{abstract}
Our goal is to populate digital environments, in which digital humans have diverse body shapes, move perpetually, and have plausible body-scene contact. 
The core challenge is to generate realistic, controllable, and infinitely long motions for diverse 3D bodies. 
To this end, we propose generative motion primitives via body surface markers, or GAMMA in short. 
In our solution, we decompose the long-term motion into a time sequence of motion primitives. We exploit body surface markers and conditional variational autoencoder to model each motion primitive, and generate long-term motion by implementing the generative model recursively.
To control the motion to reach a goal, we apply a policy network to explore the generative model's latent space and use a tree-based search to preserve the motion quality during testing.
Experiments show that our method can produce more realistic and controllable motion than state-of-the-art data-driven methods. 
With conventional path-finding algorithms, the generated human bodies can realistically move long distances for a long period of time in the scene. Code is released for research purposes at: \url{https://yz-cnsdqz.github.io/eigenmotion/GAMMA/}
\end{abstract}

\section{Introduction}
\label{sec:intro}
In recent years, the rapid development of 3D technologies has accelerated the creation of a digital replica of the real world and initiated new ways that people interact with the world and communicate with each other. 
However, there is no existing solution to automatically populate the digital world with realistic virtual humans, which move and act like real ones. 
This work aims to enable virtual humans to cruise within a 3D digital environment, similar to Odysseus, who arrived home after wandering and hazards. 
The virtual humans follow randomized routes, pass individual waypoints, and reach the destination, while retaining realistic body shape, pose, and body-scene contact.
Such technology can considerably enrich AR/VR user experiences and has many downstream applications.
For example, having virtual humans strolling inside the digital model of a medieval city can make the experience more vivid, which allows real users to follow their guidance for better sightseeing.
Beyond AR/VR, virtual humans can provide architects with a blueprint, enabling better foresight into design functionalities and defects.

This is particularly relevant to character animation in graphics, foremost in the gaming industry.
Conventionally, a set of 3D characters are pre-designed, and a motion dataset is pre-recorded. To let characters respond to user inputs or background events, motions from the dataset are created via motion graph~\cite{kovar2008motion} or motion matching~\cite{buttner2015motion}.
Although current AAA games demonstrate highly realistic character motion, this conventional technology cannot easily handle a massive number of characters with different behaviors~\cite{holden2018character}. Motion generalization across diverse characters usually requires extra motion re-targeting procedures~\cite{aberman2020skeleton}.
Moreover, the generated motions are often deterministic, close to the pre-recorded clips, and hence have limited diversity.

The availability of large-scale motion capture datasets (e.g.~AMASS~\cite{AMASS:ICCV:2019}) facilitates the learning of generative motion models. 
They can effectively produce motions based on the motion in the past~\cite{zhang2021we}, action labels~\cite{petrovich21actor}, scene context~\cite{hassan_samp_2021} and music~\cite{li2021learn}, without limit on a specific character. 
Although the motion realism is improved by replacing 3D skeletons with expressive body meshes, e.g., SMPL-X~\cite{SMPL-X:2019}, the generated motion is limited to a few seconds. It often has jittering, foot-skating, and other issues. 
To populate the digital environment,  we need a fully automated way to generate long-term (potentially infinite) and realistic motions for a large variety of human shapes.

This is a considerably challenging task and far beyond the scope of existing solutions to our knowledge.
The first obstacle we encounter is how to generate infinitely long, diverse, and stochastic human motion sequences.
Existing methods regard motion as a standard time sequence of high-dimensional feature vectors and propose to model it with a single deep neural network.
However, the uncertainty of human motion grows as time progresses. It is unclear whether a deep neural network has sufficient power to represent a perpetual motion.

To overcome this issue, we decompose a long-term motion into a time sequence of motion primitives, model each primitive, and compose them to obtain a long-term motion.
Our insight comes from psychological studies.
Human reaction time to visual stimuli is about 0.25 seconds~\cite{rt1, teichner1954recent}. 
Namely, humans cannot control their body motion immediately after seeing a signal but have to wait for ~0.25 seconds to give a response due to body inertia.
Therefore, we let a motion primitive span 0.25 seconds.
In this case, it mainly contains unconscious body dynamics, which are shorter, more deterministic, and easier to model. 
Specifically, we exploit the body surface markers to represent the body in motion~\cite{zhang2021we} and use conditional variational autoencoder (CVAE)~\cite{kingma2013auto,sohn2015learning} to model body dynamics. 
To efficiently recover the 3D body from markers, we design a body regressor with recursions.
By blending the marker predictor and body regressor, our model can synthesize realistic long-term motion, which is perceptually similar to high-quality mocap sequences, e.g.~from AMASS~\cite{AMASS:ICCV:2019}.
Of note, our marker-based motion primitive is generalizable to various body shapes, which initiates populating the 3D scene with a massive amount of virtual humans of different identities.

The second challenge is how to let the virtual humans move naturally within 3D scenes, towards a designated destination, while considering the geometric constraints of the environment.
Inspired by Ling et al.~\cite{ling2020character}, we propose a novel motion synthesis pipeline with control, which consists of a policy network and a tree-based search mechanism.
We formulate long-term motion generation as a Markov decision process, and use a policy network to explore the CVAE latent space. By sampling from the policy, the body can gradually move to the goal, while keeping the foot-ground contact plausible.
Simultaneously, we organize the motion generation process into a tree structure, which searches best motion primitives at each generation step and rejects unrealistic ones.
We perform experiments to evaluate motion realism and controllability. 
Results show that our method can produce realistic long-term motions in 3D scenes and outperform state-of-the-art methods.
Combing with conventional path-finding algorithms, e.g., navigation mesh baking and A* search~\cite{hart1968formal,snook2000simplified}, we can populate large-scale 3D scenes with a massive number of virtual humans, which have diverse body shapes, cruise following paths, and finally reach their destinations.

We name our method {\bf GAMMA}, for  \textbf{G}ener\textbf{A}tive \textbf{M}otion primitive via body surface \textbf{MA}rkers. 
Code and model are released for research purposes.

\section{Related Work}
\label{sec:related}

\myparagraph{Character animation.}
A 3D character is normally rigged with a skeleton. To animate it, mocap data is used to actuate the skeleton, and the body mesh is deformed accordingly. 
To control the character, e.g.~following a walking path, methods like motion graph~\cite{kovar2008motion} or motion matching~\cite{buttner2015motion,zinno2019ml,buttner2019machine} are to search for the most suitable motion clips from a dataset and blend them to remove discontinuities. 
Despite high motion realism, these methods are not well scalable to animate a massive number of characters with a vast mocap dataset~\cite{holden2018character}.
Although recent methods like~\cite{holden2020learned,lee2021learning} exploit neural networks to improve scalability and efficiency, the generated motions are often deterministic, close to the training samples, and hence lack diversity.

\myparagraph{Generative models for 3D body motion synthesis.}
Based on large-scale datasets, learned generative models can synthesize motions based on certain conditions.
For motion prediction, the model aims at generating motion that is expected to be close to the ground truth, provided on a motion from the past~\cite{aksan2020attention,aksan2019structured,cai2020learning,cui2020learning,ghosh2017learning,gopalakrishnan2019neural,gui2018adversarial,li2018convolutional,li2020dynamic,mao2019learning,martinez2017human,pavllo2019modeling,tang2018long,wei2020his,zhou2018autoconditioned,dang2021msr,liu2021motion}. 
When considering motion uncertainty, stochastic motion prediction is proposed to generate diverse plausible motions based on the same motion seed~\cite{barsoum2018hp,bhattacharyya2018accurate,dilokthanakul2016deep,ling2020character,walker2017pose,yan2018mt,yuan2020dlow,mao2021generating,aliakbarian2021contextually}.
Although these methods can generalize well across various body shapes and actions, motion realism is seldom considered. For example, the body is represented with stick figures. The generated motion has a limited time horizon. Scene constraints are not considered, global body configurations are not included.

More recent works exploit parametric body models e.g.~SMPL-X~\cite{SMPL-X:2019}, and directly produce motions of expressive body meshes. 
Compared to the skeleton, the motion of 3D bodies looks more realistic to our human eye and is directly compatible with existing rendering pipelines.
Zhang et al.~\cite{zhang2021we} extend stochastic motion prediction from stick figures to 3D bodies, which generate diverse global motions using a body surface marker-based representation.
Petrovich et al.~\cite{petrovich21actor} proposes a transformer VAE to generate stochastic 3D body motions based on action labels.
Zhang et al.~\cite{zhang2020perpetual} model dynamics of SMPL-X~\cite{SMPL-X:2019} body parameters with a recurrent network for generating perpetual motions of diverse 3D bodies.
Wang et al.~\cite{wang2021synthesizing} generate scene-aware root motion trajectories and synthesize body poses along the trajectory. 
Test-time optimization procedures are then applied to improve the body-scene contact.
Stark et al.~\cite{starke2019neural} propose a neural state machine model to synthesize character motions interacting with objects, such as sitting on the chair and carrying the box. 
Hassan et al.~\cite{hassan_samp_2021} propose to generate stochastic and dynamic body-scene interactions by synthesizing the goal and the motion individually with neural networks. 
Ling et al.~\cite{ling2020character} design an autoregressive VAE to model body dynamics of two consecutive frames, and exploit reinforcement learning to train a policy for task-driven motion synthesis.
Rempe et al.~\cite{rempe2021humor} propose an autoregressive model to learn 3D body motions, which is generic for body shapes and motion types. Although it focuses on motion estimation, HuMoR can generate random and long-term motions with plausible body-ground contact.

\myparagraph{Body motion control.}
Motion can be controlled in different ways. 
A straightforward approach is to extend a generative motion model by introducing control signals as additional conditions, as in~\cite{pavllo2018quaternet,habibie2017recurrent,holden2017phase,kania2021trajevae}.
Despite their effectiveness, the learned model is sensitive to the train-test domain gap, generating invalid motions when the control signal is considerably different from the training data.
Another control approach is optimization-based, in which the control signal, e.g., fixed foot position when contacting with the ground, is regarded as the data term, and the generative model is used for regularization, e.g.~\cite{wang2019combining,holden2016deep}. This shares similarity with body motion recovery from observations as in~\cite{rempe2021humor,zhang2021learning,holden2015learning,liu20204d}. They can produce high-quality results, but have high computational costs.

Many reinforcement learning (RL) methods are designed to control infinitely long motion to complete specific tasks. 
Most of them exploit physical simulation for body dynamics modeling, and propose policy networks to control, e.g. joint torques.
To obtain human-like motions, additional data, e.g. videos or mocap sequences, is used for the character to imitate~\cite{2018-TOG-deepMimic,peng2018sfv,merel2018neural,bergamin2019drecon}.
To automatize which motion to imitate, Peng et al.~\cite{peng2021amp} propose an adversarial imitation learning mechanism, designing an adversarial motion style reward during policy training.
Yuan et al.~\cite{yuan2021simpoe} focus on physics-aware motion estimation. They create a humanoid according to the SMPL blend skinning, and thus their humanoid is more human-like than the skeleton. 
Ling et al.~\cite{ling2020character} first propose a CVAE-based generative motion model, and then train a policy to produce latent variables for motion control. 

\myparagraph{Ours vs. others.}
Our method is data-driven, so no need to create a humanoid in physical simulation as in~\cite{merel2018neural,yuan2021simpoe}.

We are inspired by MotionVAE~\cite{ling2020character} and HuMoR~\cite{rempe2021humor} for synthesizing task-oriented behavior and motion estimation, respectively, in which the generative model's latent variables are manipulated to produce desired motions. 
Similar to MotionVAE~\cite{ling2020character}, our method contains a generative motion model and a policy network, but is extended with novel motion models and control methods.
Specifically, we first learn models of marker-based motion primitives, using the large-scale motion dataset AMASS~\cite{AMASS:ICCV:2019}, and then learn the policy based on the pre-trained motion model, so our method is generic for various body shapes and action types.
The policy output distribution is regularized by a KLD term in addition to the PPO term (see Eq.~\eqref{eq:policy_loss}), to encourage the motion naturalness.
Moreover, we exploit a tree-based search scheme to further improve the motion quality during testing.
Therefore, our method is suitable for populating 3D scenes in an efficient and scalable manner.



\section{Method}
\label{sec:method_motion_gen}

\subsection{Preliminaries}

\myparagraph{SMPL-X~\cite{SMPL-X:2019}.}
SMPL-X is a differentiable parametric body model.
Provided a compact set of body parameters, it yields an expressive body mesh of a fixed topology with 10,475 vertices, including face and hand details.
In our work, the body parameter set is denoted as $\bm{B} = \{\bm{\beta}, \bm{\Theta} \}$. The body shape $\bm{\beta}\in\mathbb{R}^{10}$ is the lowest 10 components in the SMPL-X body shape PCA space. 
The body configuration has $\bm{\Theta} = \{\bm{t}\in\mathbb{R}^3, \bm{R}\in\mathbb{R}^3, \bm{\theta}\in\mathbb{R}^{63}, \bm{\theta_h}\in\mathbb{R}^{24} \}$, including root translation, root orientation in axis-angle, 21 joint rotations in axis-angle, and two hand poses in the MANO PCA space~\cite{MANO:SIGGRAPHASIA:2017}, respectively. We then denote a SMPL-X mesh as $\mathcal{M}(\bm{B})$.
By sampling ${\bm \beta}$ from the standard normal distribution and sampling ${\bm \theta}$ from the Vposer~\cite{SMPL-X:2019}, we can obtain 3D bodies of diverse shapes and poses quickly.
Although our method is implemented with SMPL-X, it can be straightforwardly extended to other parametric body models, ~e.g.~\cite{STAR:ECCV:2020,xu2020ghum}.

\myparagraph{MOJO~\cite{zhang2021we}.}
MOJO is a solution to stochastic motion prediction of 3D body meshes, including a surface marker-based representation, a CVAE motion model, and a recursive scheme for 3D body recovery.
Corresponding to the marker placement in mocap systems, the MOJO body markers are selected from the SMPL-X mesh template, converting the body motion into a time sequence of point clouds.
Compared to the representations like joint locations, surface markers have richer body shape information and provide more constraints on the body degrees-of-freedom (DoF).
The reprojection scheme exploits these benefits and effectively recovers the body mesh from predicted markers via optimization. At each time step during inference, markers are selected on the recovered body mesh, and then used as input for the next time step. Since the markers are reprojected to the valid body shape space repeatedly, the marker prediction error is hardly accumulated as time progresses, which can therefore retain the 3D body valid.

Despite several advantages, MOJO produces motions with limited length, and cannot control the motion generation process.
In addition, the body fitting optimization significantly increases the computation time.

\myparagraph{Motion formulation.}
We decompose a long-term motion into a time sequence of motion primitives with overlaps. 
Each motion primitive contains a motion seed ${\bm X}=\{{\bm x}_1,...,{\bm x}_M\}$ with $M=\{1,2\}$, denoting the overlapped frames, and the future frames ${\bm Y}=\{{\bm y}_1,...,{\bm y}_N\}$. 
Each frame is represented by the concatenation of the 3D marker locations.
The motion primitive is defined in a canonical space. 
Following MOJO~\cite{zhang2021we}, the canonical coordinate is located at the pelvis in the frame ${\bm x}_1$. The X-axis is the horizontal component pointing from the left hip to the right hip, the Z-axis is pointing up, and the Y-axis is pointing forward.

\begin{figure}[t]
    \centering
    \includegraphics[width=\linewidth]{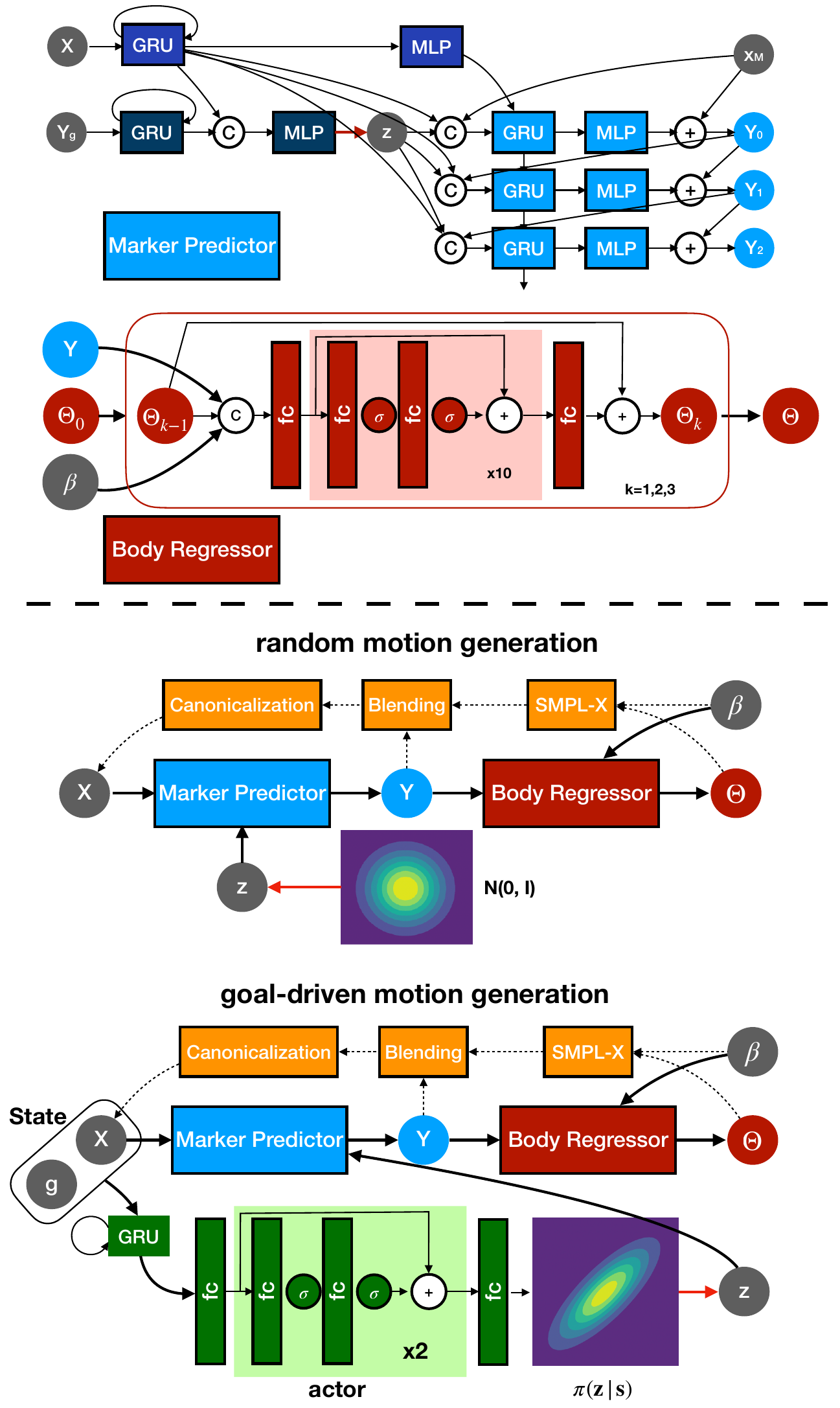}
    \caption{GAMMA architectures. The first two diagrams illustrate the marker predictor and the body regressor, respectively. The bottom two show their combinations. Random motion can be synthesized via sampling ${\bm z}$ from the standard normal distribution, and goal-driven motion can be synthesized via sampling ${\bm z}$ from the policy output. The red arrows denote the sampling operation, and the dash curves denote implementing motion primitive models recursively. See respective sections for detailed demonstrations. }
    \label{fig:network}
    \vspace{-3mm}
\end{figure}

\subsection{Generative Motion Primitives}
\label{sec:motion_primitives}

\subsubsection{Model Architecture}
The network architectures of GAMMA are illustrated in Fig.~\ref{fig:network}.
The generative motion primitive model consists of a marker predictor and a body regressor. 

\myparagraph{The marker predictor.}
In our work, we exploit the `SSM2 67' marker placement in MOJO~\cite{zhang2021we} due to its better empirical performance.
The marker predictor is cast by a CVAE, which has a condition branch, an encoder, and a decoder. The encoder is only used during training. 
During inference, we can randomly draw ${\bm z}$ from $\mathcal{N}(0, {\bm I})$ to obtain different ${\bm Y}$ based on the same motion seed ${\bm X}$. 
Compared to~\cite{yuan2020dlow, zhang2021we}, our model has no additional sampling module and latent DCT, and hence is easier to train.
Depending on whether the motion seed ${\bm X}$ contains body dynamics, we design a 1-frame predictor and a 2-frame predictor, denoting the number of frames in the motion seed ${\bm X}$. 
Their properties are investigated in the appendix.

\myparagraph{The body regressor.}
This network learns to recover the global translations and the joint rotations simultaneously from the markers within a motion primitive. 
It takes the predicted markers, an initial body parameter $\Theta_0$, and a provided body shape as input.
Similar to HMR~\cite{kanazawa2018end}, the residual blocks are employed in a recursive manner.
We set $\Theta_0$, including the global translation, orientation, body pose, and hand pose, to zero in our implementation.
To improve back-propagation, the global body orientation and joint rotations are converted into the 6D continuous representation~\cite{zhou2019continuity} in the input and inside of the network, and changed back to axis-angle at the output.
Since the body shape is dependent on gender, we also propose two versions of body regressors for males and females, respectively, which are separately trained.

\subsubsection{Training}
\label{sec:model_training}

\myparagraph{Training the marker predictor.}
First, we train the predictor to learn each individual motion primitive.
Similar to MOJO~\cite{zhang2021we}, the training loss contains a reconstruction term and a robust KL-divergence term to regularize the latent space, which is formulated to

\begin{equation}
    \begin{split}
    \mathcal{L}_{predictor} &= \mathbb{E}_{{\bm Y}}[|{\bm Y}-{\bm Y}^{rec}|] + \lambda \mathbb{E}_{{\bm Y}}[|\Delta{\bm Y}-\Delta{\bm Y}^{rec}|] \\ &+ \Psi\left(\text{KL-div}( q({\bm Z} | {{\bm X},{\bm Y}}) || \mathcal{N}(0, {\bm I})) \right),
    \end{split}
    \label{eq:trainloss}
\end{equation}
in which $^{rec}$ means the reconstructed variable, $\Delta$ denotes the time difference, and $\lambda=3$ in our implementation.
The KLD term employs a robust function $\Psi(s) = \sqrt{1+s^2}-1$~\cite{charbonnier1994two}, which automatically reduces the gradient of the KLD term, when its value is small, so as to alleviate posterior collapse.

Second, we fine-tune the predictor by rolling out longer sequences, as in~\cite{martinez2017human,ling2020character,rempe2021humor}. 
Specifically, we use the last one or two frames of a generated motion primitive as the motion seed to generate the next motion primitive, and we minimize the same loss Eq.~\eqref{eq:trainloss} as above. The transformation between motion primitives is based on the ground truth canonical coordinate.
Since this rollout training process takes prediction errors into account, learned generative models can produce long motions stably during testing, and favors recovering body shape when the motion seed is not fully valid.
In our experiments, the time horizon of the rollout is set to 8 motion primitives. The 1-frame model and the 2-frame model are trained separately.

\myparagraph{Training the body regressor.}
The body regressor is trained with batches of canonicalized motion primitives. 
The training loss is based on forward kinematics, and is given by
\begin{equation}
    \mathcal{L}_{regressor} = |M \circ \mathcal{M}({\bm \Theta}, {\bm \beta})-V_{gt}| + \alpha |{\bm \theta_h}|^2,
    \vspace{-1mm}
\end{equation}
in which $M\circ$ denotes selecting marker vertices from the mesh template, $V_{gt}$ denotes the ground truth body surface markers, and $\alpha=0.01$ in our work. This hand regularization term is necessary, since there are only 3-4 markers for each hand, and none of them are on the fingers.
The regressor is trained for male and female bodies separately.
Although we only train the body regressors with ground truth markers, it performs stably when taking predicted markers as input in our trials. We did not observe the advantages of training the predictor and the regressor jointly in our experiments (see Sec.~\ref{sec:experiment:exp1}).

\subsection{Motion Synthesis and Control}
\label{sec:motion_control}

By implementing GAMMA recursively, 
we can either synthesize random motion by sampling ${\bm z}$ from $\mathcal{N}(0, {\bm I})$ at all steps, or control the motion to reach the goal by additionally applying a policy network. 
The recursion mechanism incorporates a SMPL-X body model, a motion blending module, and a canonicalization module, which are illustrated by dash curves and orange blocks in Fig.~\ref{fig:network}.

The SMPL-X module is to obtain the markers on the body mesh, which is regressed from predicted markers. Compared to the optimization-based reprojection scheme in MOJO~\cite{zhang2021we}, our regression is significantly faster without degrading the accuracy (see Tab.~\ref{tab:regressor_test}).
However, due to lack of regularization, the regressor cannot guarantee the body is always valid. 
Invalid bodies can produce corrupted motion seeds, and hence destroy all future motion generation steps.
Therefore, the blending module is essential to keep the recursion stable.
Although one can use the predicted markers as the motion seed only, prediction error gradually accumulates over time in this case, causing high-frequent body movements and jitters.

\subsubsection{Goal-driven Motion Policy}

Inspired by~\cite{ling2020character}, we formulate motion synthesis as a Markov decision process and control it via RL to reach a goal. 
Under this setting, we define a time sequence of tuples $\{\left({\bm s}_t, {\bm a}_t, r_t \right)\}_{t=0}^{T}$, denoting the state, the action, and the reward at each primitive generation step.

\myparagraph{The state.}
The state incorporates the motion seed and the normalized vectors from individual markers pointing to the goal in the canonical space. 
Given the goal location ${\bm g}\in\mathbb{R}^{3}$ on the ground plane, the state can be given by
\begin{equation}
    \label{eq:states}
    {\bm s}_t = \left({\bm X}, ({\bm g}-{\bm X})_{\bm n}  \right)^T \in \mathbb{R}^{l \times V \times 6},
\end{equation}
in which $l \in \{1,2\}$ denotes the length of the motion seed, $V$ denotes the number of markers, and $(\cdot)_{\bm n}$ denotes the normalized 3D vector with unit length. 
This state has balanced dimensions between the motion seed and the goal-based features. Normalizing the vector length is helpful to stabilize the policy training in our trials.

\myparagraph{The action.}
Like in~\cite{ling2020character}, we regard the latent variable ${\bm z}$ as the action to generate the current motion primitive. 

\myparagraph{The reward.}
The reward evaluates the quality of a generated motion primitive, which is given by
\begin{equation}
    \label{eq:reward}
    r = r_{path} + \beta_1 r_{ori} + \beta_2 r_{contact} + \beta_3 r_{vposer} + \beta_4 r_{goal},
\end{equation}
including the path following reward, the body orientation reward, the body-ground contact reward, the valid body pose reward, and the goal-reaching reward, respectively. 
Each of them ranges from 0 to 1.
$r_{path}$ encourages moving towards the goal while following the straight path, which can be given by
\begin{equation}
    r_{path} = \frac{\langle({\bm p}^{xy}_{T} - {\bm p}^{xy}_{0})_{\bm n}, ({\bm g^{xy}} - {\bm p}^{xy}_{0})_{\bm n} \rangle +1}{2},
\end{equation}
in which $({\bm p}^{xy}_{T} - {\bm p}^{xy}_{0})_{\bm n}$ denotes the normalized pelvis movement along the ground plane, and $({\bm g^{xy}} - {\bm p}^{xy}_{0})_{\bm n}$ denotes the normalized direction to the goal along the ground plane. 
$r_{ori}$ encourages the body to face the goal and can be formulated as
\begin{equation}
    r_{ori} = \frac{\langle {\bm o}^y_{T}, ({\bm g^{xy}}-{\bm p}^{xy}_T)_{\bm n} \rangle+1}{2},
\end{equation}
in which ${\bm o}_{T}$ denotes the unit vector of the body facing direction at the last motion primitive frame. 
$r_{contact}$ encourages the body moving along the ground plane and discourage skating, which is formulated as
\begin{equation}
    r_{contact} = e^{-|{\bm p}^z_{T} - {\bm p}^z_{0}|} \cdot e^{-|{\bm Y}_{vel}^*|_2},
    \vspace{-1mm}
\end{equation}
which discourages the pelvis Z-location shift and encourages the minimal marker speed close to zero within a motion primitive.
$r_{vposer}$ encourages the body pose to keep valid, and is formulated as
\begin{equation}
    r_{vposer} = e^{-|{\bm \mu}\left({\bm \theta}\right)|_2},
\end{equation}
which gives a higher value if the encoded pose in the VPoser~\cite{SMPL-X:2019} latent space is closer to 0. The reward $r_{goal}$ is added to all the motion primitives in the entire motion sequence, according to the closest distance between any motion primitive to the goal.  
This reward can be formulated as
\begin{equation}
    r_{goal} = 
        \begin{cases}
          \left(\frac{L-|{\bm p}^{xy}_{*}-{\bm g}|_2}{L}\right)_{+} & \text{if} |{\bm p}^{xy}_{*}-{\bm g}|_2 > \epsilon\\
          1 & \text{otherwise}
        \end{cases}
        \vspace{-1mm}
\end{equation}
in which ${\bm p}^{xy}_{*}$ denotes the pelvis XY-location at the last frame of the closest motion primitive, $L$ denotes the range of the activity area,  $(\cdot)_{+}$ thresholds values to be non-negative, and $\epsilon$ is the tolerance defining the goal is reached. 

\myparagraph{Policy network and training.}
We employ the actor-critic mechanism~\cite{sutton1998introduction} for training the policy. The actor produces a diagonal Gaussian distribution $\pi({\bm z}_t | {\bm s}_t)$, and its architecture is illustrated in Fig.~\ref{fig:network}. 
The critic network shares the GRU with the actor and is only used in the training phase to estimate the expected return for the advantage function.

To train the policy, we set up a simulation environment. 
The area range $L$ is a $20\times20 m^2$ ground plane. At the beginning of the simulation, we create a character with a random gender, body shape, and pose, and place it in the center of the area with a random facing orientation. We then randomly set a goal on the ground following a uniform distribution.
We first use the 1-frame model to produce 32 motion primitives from the same initial body, and then use the 2-frame model to generate motions for each body clone. 
The simulation terminates after generating 60 primitives, or all motions reach the goal. 
The goal is regarded as reached if its horizontal distance to the body pelvis is smaller than 0.75m in our trials.

We train the policy network for 500 epochs and run the simulation 8 times for each epoch. Therefore, we collect motion data from 4000 random body-goal pairs.
We update the policy network at each epoch by minimizing the following loss 
\begin{equation}
\begin{split}
    \mathcal{L}_{policy} &= \mathcal{L}_{PPO} + \mathbb{E}\left[ (R_t - V({\bm s}_t))^2 \right] \\
    &+ \alpha \Psi\left(\text{KL-div}( \pi({\bm z} | {\bm s}) || \mathcal{N}(0, {\bm I}))) \right),
\end{split}
\label{eq:policy_loss}
\end{equation}
in which the first term updates the policy network with PPO~\cite{schulman2017proximal}, the second term is to update the critic network for better value estimation, and the third term is to regularize motion in the latent space. $R_t$ denotes the expected return from rewards with a discount factor.
More details are given in the appendix.

\subsubsection{Tree-based Search}

Given the probabilistic nature of our generative model, there is no hard constraint on the body-scene interaction, when sampling motions from the latent space. 
Therefore, we exploit a tree-based search to discard motion primitives with inferior body-ground contact during test time.

Inspired by random tree-based motion planning~\cite{lavalle1998rapidly,lavalle2001rapidly}, we organize the motion generation process into a tree structure, with the root being the initial body pose. 
Each node denotes a motion primitive, and hence has one parent and multiple children. 
In addition, each node has a quality or cost value, and all nodes at the same level are ranked.
During generation, we can keep the best $K$ nodes to yield new motions, and discard the rest.
Specifically, at the first level, we generate $N$ primitive nodes from the root, and only keep the best $K$ while discarding others. 
At the second level, we will have $KN$ nodes after generation, and we keep the best $K$. And so on. 
This tree-based search works both for ${\bm z} \sim \mathcal{N}(0, {\bm I})$ and  ${\bm z} \sim \pi({\bm z}|{\bm s})$. With the learned policy, the search space is significantly reduced. 

The design of the node quality/cost depends on the application scenario and can be different from the reward. 
For example, one can set a high weight on the foot-ground contact to discard floating bodies, or put a high weight on the distance to the goal to encourage fast movement.

\section{Experiment}
\label{sec:experiment}
In this section, we perform empirical evaluations on motion realism and motion controllability. 
Additional experiments, including the performance of the marker predictor and the body regressor, comparison with other body representations, the influence of bodies on motion generation, and runtime test, are presented in the appendix.

\myparagraph{Dataset.}
We exploit the large-scale mocap data {\bf AMASS}~\cite{AMASS:ICCV:2019}.
Specifically, we train the marker predictors on {\bf CMU}~\cite{mocap_cmu}, {\bf MPI HDM05}~\cite{muller2007documentation}, {\bf BMLmovi}~\cite{ghorbani2021movi}, {\bf KIT}~\cite{Mandery2016b} and {\bf Eyes Japan}~\cite{mocap_eyesjapan}.
When training the body regressors, we additionally use {\bf BMLrub}, {\bf Transitions} and {\bf TotalCapture}~\cite{trumble2017total}, but exclude {\bf KIT} for testing (see Sec.~\ref{sec:regressor_test}). 

For each mocap sequence, we downsample it to 40fps, and trim it into motion primitives. Each motion primitive clip contains 10 frames or 0.25 seconds, and is canonicalized before model training.
In addition, we also prepare longer canonicalized clips with 10 motion primitives, which are used for training with rollout.

\subsection{Evaluation on Motion Realism}
\label{sec:experiment:exp1}
 
We randomly select 100 static poses from {\bf HumanEva} and {\bf ACCAD}, respectively, and generate a 10-second motion based on each static pose.

\myparagraph{Baseline and our methods.}
To our knowledge, HuMoR~\cite{rempe2021humor} is the state-of-the-art generative model for generating long-term motions of diverse 3D bodies.
To produce motions with static poses like ours, we set zero velocities to the initial bodies and generate 10-second motions. 

Our method has several versions.
`ours-e2e' denotes training predictor and regressor end-to-end (see appendix).
`ours-ro' exploits a marker predictor with rollout training, and blends the predicted markers and the reprojected markers by averaging.
`ours-reproj' exploits a marker predictor without rollout training, and only uses the reprojected markers on the 3D body as the next motion seed.
The suffix `-xf' denotes that the number of frames in the motion seed is x.
We also use `ours-ro-policy' to test the influence of the policy network. In this case, we set a random goal for each initial static body, and generate a 10-second motion. 
Moreover, we randomly choose 200 sequences from {\bf AMASS} to measure how far we are from generating real motion.
For a fair comparison, neither test-time optimization nor tree-based search is performed in this experiment.

\myparagraph{Evaluation metrics.}
The metrics are about \textit{body-ground contact} and \textit{perceptual study}.
For the \textit{body-ground contact}, we set a threshold height of 0.05m from the ground plane and a speed threshold of 0.075m/s for skating. 
Then the contact score is defined as 
\begin{equation}
    \label{eq:contact_score}
    s = e^{-(\min|{\bm Y}_z|-0.05)_{+}} \cdot e^{-(\min|{\bm Y}_{vel}|-0.075)_{+} },
    \vspace{-1mm}
\end{equation}
in which $|\bm{Y}_z|$ and $|\bm{Y}_{vel}|$ denote the height absolute value and the velocity magnitude, respectively. $(\cdot)_+$ denotes the zero-threshold function equivalent to ReLU. 
Therefore, this score ranges from 0 to 1, and is the higher, the better.
For the \textit{perceptual study}, we render the 100 generated motions from {\bf HumanEva} and let Amazon Mechanical Turk users evaluate on a six-point Likert scale from ‘strongly disagree‘ (1) to ‘strongly agree’ (6), similar to ~\cite{zhang2020generating,zhang2021we}.

\begin{figure}[t]
    \centering
    \includegraphics[width=\linewidth]{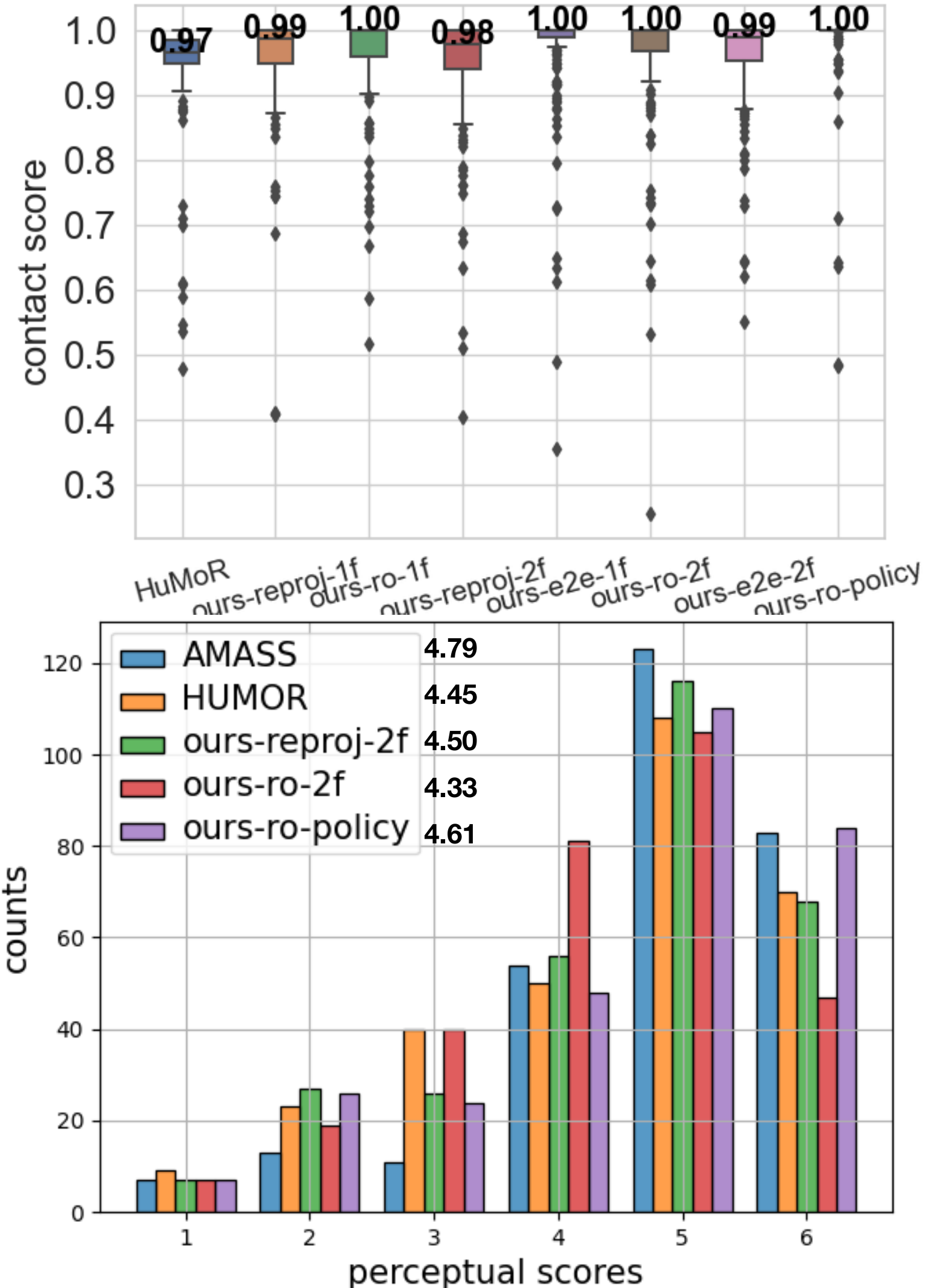}
    \vspace{-3mm}
    \caption{Motion realism analysis. From top to bottom, show the body-ground contact and the perceptual score, respectively.
    At the top, the box plot denotes the lower and the upper quartiles, and the numbers denote the median.
    At the bottom, the X and Y-axis denote the perceptual scores and the total counts, respectively. The mean perceptual scores are shown beside the legend.
    }
    \label{fig:random_gen}
\end{figure}

\myparagraph{Results and discussion.}
The results are shown in Fig.~\ref{fig:random_gen}.
First, we can see that our methods consistently outperform HuMoR w.r.t. the body-ground contact. 
In particular, the motion generated by the policy network outperforms all others.
This shows the contact reward for the policy training is effective. We can also observe that the end-to-end training is not favorable in this case.
Second, user study results show that our methods produce perceptually more realistic motions than HuMoR.
By comparing `ours-ro-policy' and `ours-ro-2f', the perceptual score increases by a large margin, indicating the policy's effectiveness. 
Since goal-driven behavior is common in our daily lives, our generated motions with goals are more perceptually realistic than long-term random motions.

According to the visualizations, we can see that the HuMoR can produce observable body jitters and other high-frequency non-plausible movements. In contrast, our methods produce smoother motions with more plausible body-ground contact.
However, we observe some bodies moving stiffly in our results, particularly in the 1-frame-based models. 
Since these methods generate motions only based on a single frame, the continuity of higher-order dynamics is not guaranteed. 
The 2-frame-based models can alleviate this issue, especially when the body motion is slow and low-frequent.

\begin{table}
    \centering
    \scriptsize
    \begin{tabular}{lccccc}
    \toprule
         & \textit{steps}$\downarrow$ & \textit{success rate}$\uparrow$ & \textit{avg. dist.}$\downarrow$ & \textit{contact score}$\uparrow$ \\
         \midrule
        g-CVAE & 24.1 & 0.91 & 3.46& 0.8 \\
        \midrule
        policy(10) & 33.9 & 0.95 & 3.32 & 0.94 \\
        policy(1) & \textbf{21.2}& \textbf{1.0}& 3.60& 0.93 \\
        policy(10) \& top1 & 26.5 & \textcolor{blue}{0.99}& \textbf{0.04}& \textbf{0.97} \\
        policy(10) \& top4 & \textcolor{blue}{23.4} & 0.96& \textcolor{blue}{1.98}& \textcolor{blue}{0.96} \\
        \bottomrule
    \end{tabular}
    \vspace{-3mm}
    \caption{Comparison between motion control methods. The up/down arrows denote the score is the higher/lower the better. Best results are in bold, second best in blue.}
    \label{tab:motion_control}
    \vspace{-3mm}
\end{table}

\subsection{Evaluation on Motion Control}
\label{sec:experiment:exp2}

\myparagraph{Datasets and evaluation metrics.}
For testing, we randomly select 100 character-goal pairs from the simulation area. 
Like policy training, each body yields 32 motions.
The simulation terminates when one body clone reaches the goal or after generating 60 primitives.

The evaluation metrics include: 
1) {\em steps}, i.e. the minimum number of motion primitives to achieve the goal. 
2) \textit{Success rate}. We regard a test as successful if anybody's clone reaches the goal within 60 motion primitives.
3) \textit{Average distance}. We average the distances of all bodies to the target at the end of each simulation, which also measures how bodies are scattered when the goal is reached.
4) \textit{Contact score}, which is defined in Eq.~\eqref{eq:contact_score}.

\myparagraph{Baseline and our methods.}
A conventional baseline is to incorporate the goal into the generative motion model as an additional condition, like in~\cite{hassan_samp_2021,starke2019neural}.
For a fair comparison, we add the goal-based features in Eq.~\eqref{eq:states} to our models and denote this baseline as `g-CVAE'.

About our methods, we use `policy ($\alpha$)' to denote the version with the KLD term weight in Eq.~\eqref{eq:policy_loss}. 
In addition, we use `top $k$' to denote selecting the best $k$ primitives at each tree level, according to the contact score and the distance to the goal.

\begin{figure}[t]
    \centering
    \includegraphics[width=0.9\linewidth]{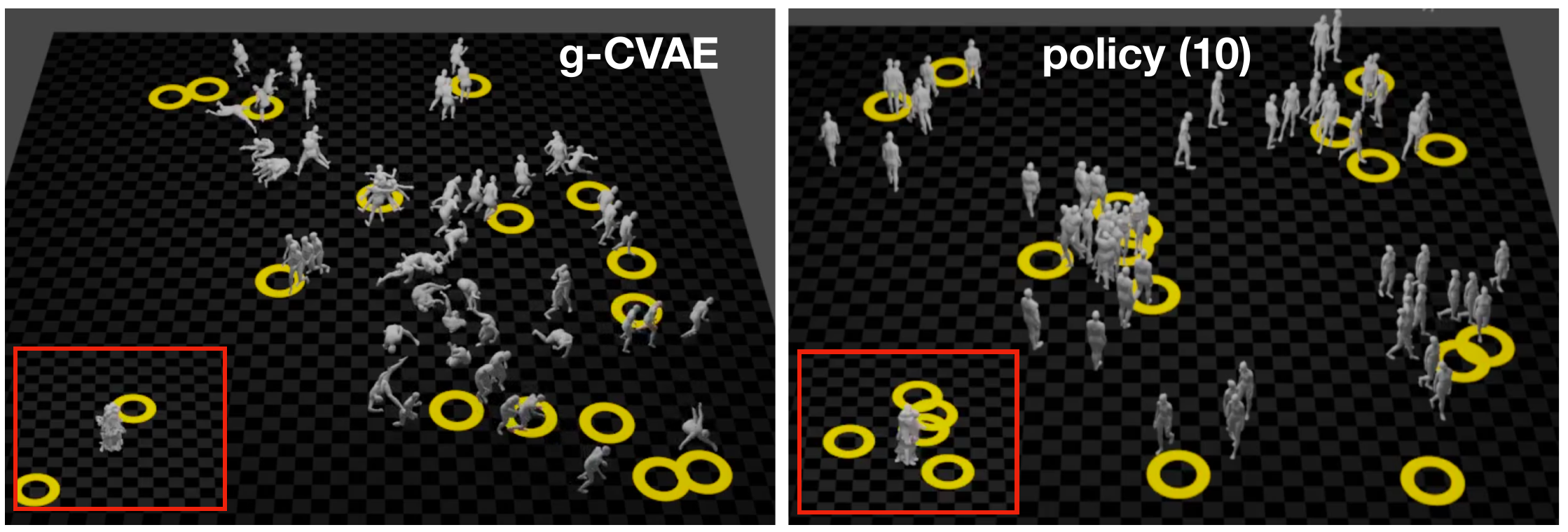}
    \vspace{-3mm}
    \caption{Perceptual comparison between control methods. 
    The subfigures and the main figures denote the starting and the ending frame, respectively. The yellow ring denotes the goal.}
    \label{fig:policy_compare}
\end{figure}

\begin{figure}[t]
    \centering
    \includegraphics[width=0.9\linewidth]{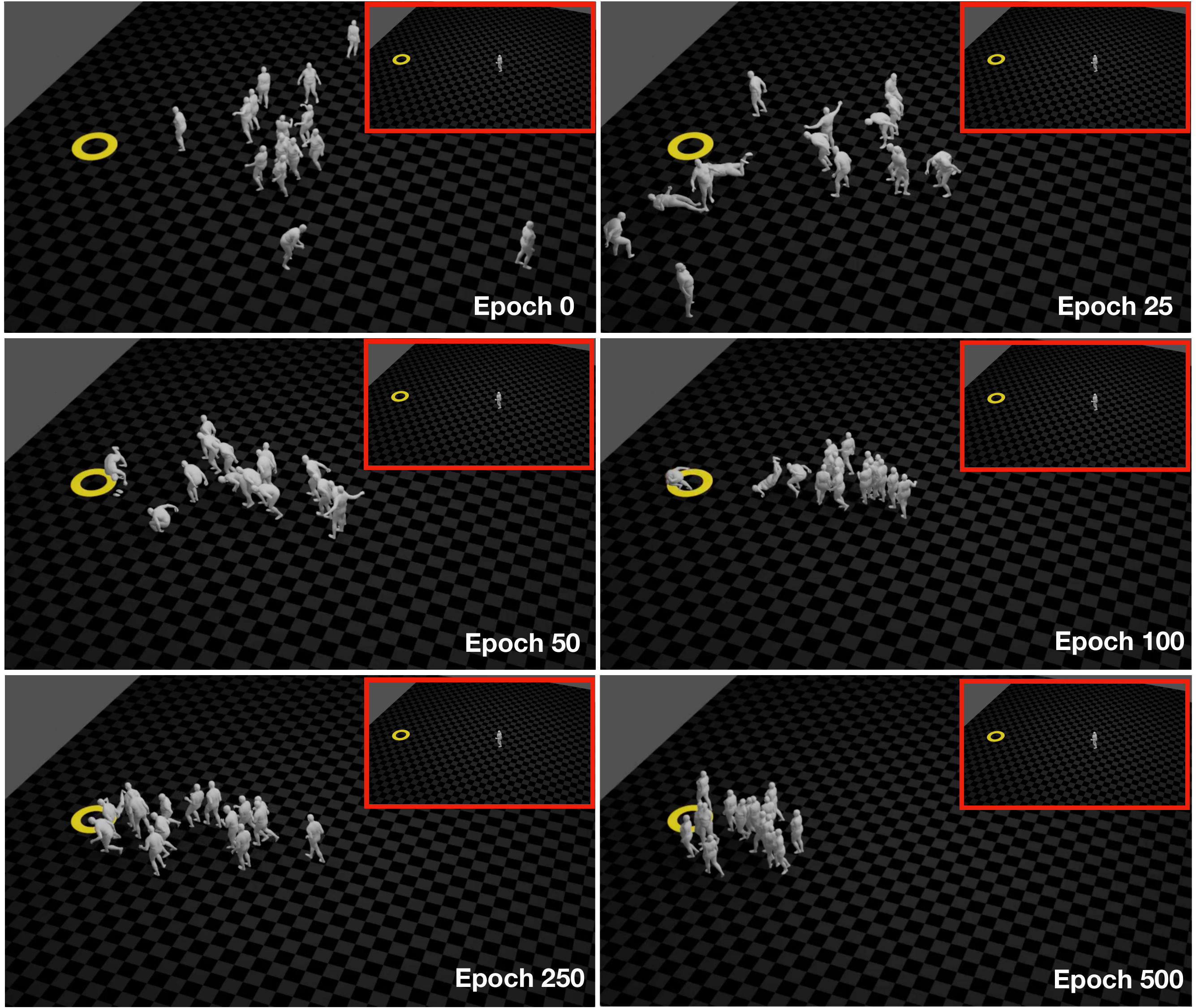}
    \vspace{-3mm}
    \caption{Visualization of policy training process. The notations are same to Fig~\ref{fig:policy_compare}, and the text indicates the training epochs.}
    \label{fig:policy_evolving}
    \vspace{-3mm}
\end{figure}

\myparagraph{Results and discussions.}
As shown in Tab.~\ref{tab:motion_control}, our policy-based methods are superior to `g-CVAE' w.r.t. all metrics, in particular the contact score.
Between our policy-based methods, we find a lower KLD weight can lead to faster but less plausible motions. When exploiting search, the performance consistently improves, and is not sensitive to the value of $k$.
Fig.~\ref{fig:policy_compare} visualizes their results. 
We find `g-CVAE' cannot keep the body valid, probably because the goals for training can only be from the data, but goals for testing are random.
On the other hand, `policy (10)' can largely preserve the motion realism while driving the bodies to the goal.
Note that `top 4' includes sub-optimal motions when calculating the average, so its number is higher than `top 1'.

Moreover, we visualize how the policy is trained in Fig.~\ref{fig:policy_evolving}. 
The initial policy network (at epoch 0) can produce valid goal-agnostic and random motions. 
In the early stage, the policy drives the body to the goal as quickly as possible, ignoring motion realism. 
With more training epochs, the KLD term in Eq.~\eqref{eq:policy_loss} provides more motion regularization. 
At 500 epochs, the goal-reaching behavior and the motion realism are well balanced. 
With more training epochs, the KLD term decreases, causing the motions to become more regularized while ignoring the goals.



\section{Conclusion}
\label{sec:conclusion}

In this paper, we propose an automatic solution to populate 3D scenes with diverse moving bodies. 
We learn generative models of body surface marker-based motion primitives and synthesize long-term motion with a policy network and tree-based search.
Experiments show the effectiveness and advantages over baselines.
Together with conventional path-finding algorithms, we can generate diverse people wandering in the digital environment.

\myparagraph{Limitations.}
The generated motion is not fully physically plausible since our method is purely data-driven.
For example, the body can tilt, ignoring gravity.
Compared to {AMASS} sequences, motion realism still has room to improve. 
Motion generated by the policy may take a long time to reach a goal in the near distance, which is different from real human behavior. 
In the future, we will extend our work to synthesize more complex body-scene interactions.

\myparagraph{Social impact.}
Although training male and female models separately can improve motion realism, our method could be potentially biased, if the male and female motion sequences are not well balanced in the dataset. 

\myparagraph{Acknowledgement.}
We appreciate Gramazio Kohler Research for providing architecture CAD models. Body visualizations are based on the SMPL-X blender add-on. This work is supported by the SNF grant 200021\_204840.

{\small
\bibliographystyle{ieee_fullname}
\bibliography{egbib}

\begin{thebibliography}{10}\itemsep=-1pt

\bibitem{rt1}
Experiment: How fast your brain reacts to stimuli.

\bibitem{mocap_eyesjapan}
{Eyes Japan} motion capture dataset.
\newblock \url{http://mocapdata.com/index.html}.

\bibitem{mocap_cmu}
{CMU} graphics lab. {CMU} graphics lab motion capture database.
\newblock \url{http://mocap.cs.cmu.edu/}, 2000.

\bibitem{accad}
{ACCAD MoCap System and Data}.
\newblock \url{https://accad.osu.edu/research/motion-lab/systemdata}, 2018.

\bibitem{aberman2020skeleton}
Kfir Aberman, Peizhuo Li, Dani Lischinski, Olga Sorkine-Hornung, Daniel
  Cohen-Or, and Baoquan Chen.
\newblock Skeleton-aware networks for deep motion retargeting.
\newblock {\em ACM Transactions on Graphics (TOG)}, 39(4):62--1, 2020.

\bibitem{aksan2020attention}
Emre Aksan, Peng Cao, Manuel Kaufmann, and Otmar Hilliges.
\newblock Attention, please: A spatio-temporal transformer for {3D} human
  motion prediction.
\newblock {\em arXiv preprint arXiv:2004.08692}, 2020.

\bibitem{aksan2019structured}
Emre Aksan, Manuel Kaufmann, and Otmar Hilliges.
\newblock Structured prediction helps {3D} human motion modelling.
\newblock In {\em IEEE Conference on Computer Vision and Pattern Recognition},
  pages 7144--7153, 2019.

\bibitem{aliakbarian2021contextually}
Sadegh Aliakbarian, Fatemeh Saleh, Lars Petersson, Stephen Gould, and Mathieu
  Salzmann.
\newblock Contextually plausible and diverse 3d human motion prediction.
\newblock In {\em Proceedings of the IEEE/CVF International Conference on
  Computer Vision}, pages 11333--11342, 2021.

\bibitem{barsoum2018hp}
Emad Barsoum, John Kender, and Zicheng Liu.
\newblock {HP-GAN}: Probabilistic {3D} human motion prediction via {GAN}.
\newblock In {\em IEEE Conf. Comput. Vis. Pattern Recog. Worksh.}, pages
  1418--1427, 2018.

\bibitem{bergamin2019drecon}
Kevin Bergamin, Simon Clavet, Daniel Holden, and James~Richard Forbes.
\newblock Drecon: data-driven responsive control of physics-based characters.
\newblock {\em ACM Transactions On Graphics (TOG)}, 38(6):1--11, 2019.

\bibitem{bhattacharyya2018accurate}
Apratim Bhattacharyya, Bernt Schiele, and Mario Fritz.
\newblock Accurate and diverse sampling of sequences based on a “best of
  many” sample objective.
\newblock In {\em IEEE Conference on Computer Vision and Pattern Recognition},
  pages 8485--8493, 2018.

\bibitem{buttner2019machine}
Michael Buttner.
\newblock Machine learning for motion synthesis and character control in games.
\newblock {\em Proc. of I3D 2019}, 2019.

\bibitem{buttner2015motion}
Michael B{\"u}ttner and Simon Clavet.
\newblock Motion matching-the road to next gen animation.
\newblock {\em Proc. of Nucl. ai}, 2015, 2015.

\bibitem{cai2020learning}
Yujun Cai, Lin Huang, Yiwei Wang, Tat-Jen Cham, Jianfei Cai, Junsong Yuan, Jun
  Liu, Xu Yang, Yiheng Zhu, Xiaohui Shen, et~al.
\newblock Learning progressive joint propagation for human motion prediction.
\newblock In {\em European Conference on Computer Vision}, 2020.

\bibitem{charbonnier1994two}
Pierre Charbonnier, Laure Blanc-Feraud, Gilles Aubert, and Michel Barlaud.
\newblock Two deterministic half-quadratic regularization algorithms for
  computed imaging.
\newblock In {\em Proceedings of the International Conference on Image
  Processing}, pages 168--172, 1994.

\bibitem{cui2020learning}
Qiongjie Cui, Huaijiang Sun, and Fei Yang.
\newblock Learning dynamic relationships for {3D} human motion prediction.
\newblock In {\em IEEE Conference on Computer Vision and Pattern Recognition},
  pages 6519--6527, 2020.

\bibitem{dang2021msr}
Lingwei Dang, Yongwei Nie, Chengjiang Long, Qing Zhang, and Guiqing Li.
\newblock Msr-gcn: Multi-scale residual graph convolution networks for human
  motion prediction.
\newblock In {\em Proceedings of the IEEE/CVF International Conference on
  Computer Vision}, pages 11467--11476, 2021.

\bibitem{dilokthanakul2016deep}
Nat Dilokthanakul, Pedro~AM Mediano, Marta Garnelo, Matthew~CH Lee, Hugh
  Salimbeni, Kai Arulkumaran, and Murray Shanahan.
\newblock Deep unsupervised clustering with gaussian mixture variational
  autoencoders.
\newblock {\em arXiv preprint arXiv:1611.02648}, 2016.

\bibitem{ghorbani2021movi}
Saeed Ghorbani, Kimia Mahdaviani, Anne Thaler, Konrad Kording, Douglas~James
  Cook, Gunnar Blohm, and Nikolaus~F Troje.
\newblock Movi: A large multi-purpose human motion and video dataset.
\newblock {\em Plos one}, 16(6):e0253157, 2021.

\bibitem{ghosh2017learning}
Partha Ghosh, Jie Song, Emre Aksan, and Otmar Hilliges.
\newblock Learning human motion models for long-term predictions.
\newblock In {\em International Conference on 3D Vision}, pages 458--466. IEEE,
  2017.

\bibitem{gopalakrishnan2019neural}
Anand Gopalakrishnan, Ankur Mali, Dan Kifer, Lee Giles, and Alexander~G
  Ororbia.
\newblock A neural temporal model for human motion prediction.
\newblock In {\em IEEE Conference on Computer Vision and Pattern Recognition},
  pages 12116--12125, 2019.

\bibitem{gui2018adversarial}
Liang-Yan Gui, Yu-Xiong Wang, Xiaodan Liang, and Jos{\'e}~MF Moura.
\newblock Adversarial geometry-aware human motion prediction.
\newblock In {\em European Conference on Computer Vision}, pages 786--803,
  2018.

\bibitem{habibie2017recurrent}
Ikhsanul Habibie, Daniel Holden, Jonathan Schwarz, Joe Yearsley, and Taku
  Komura.
\newblock A recurrent variational autoencoder for human motion synthesis.
\newblock In {\em 28th British Machine Vision Conference}, 2017.

\bibitem{hart1968formal}
Peter~E Hart, Nils~J Nilsson, and Bertram Raphael.
\newblock A formal basis for the heuristic determination of minimum cost paths.
\newblock {\em IEEE transactions on Systems Science and Cybernetics},
  4(2):100--107, 1968.

\bibitem{hassan_samp_2021}
Mohamed Hassan, Duygu Ceylan, Ruben Villegas, Jun Saito, Jimei Yang, Yi Zhou,
  and Michael Black.
\newblock Stochastic scene-aware motion prediction.
\newblock In {\em Proc. International Conference on Computer Vision (ICCV)},
  pages 11374--11384, Oct. 2021.

\bibitem{holden2018character}
Daniel Holden.
\newblock Character control with neural networks and machine learning.
\newblock {\em Proc. of GDC 2018}, 2018.

\bibitem{holden2020learned}
Daniel Holden, Oussama Kanoun, Maksym Perepichka, and Tiberiu Popa.
\newblock Learned motion matching.
\newblock {\em ACM Transactions on Graphics (TOG)}, 39(4):53--1, 2020.

\bibitem{holden2017phase}
Daniel Holden, Taku Komura, and Jun Saito.
\newblock Phase-functioned neural networks for character control.
\newblock {\em ACM Transactions on Graphics (TOG)}, 36(4):1--13, 2017.

\bibitem{holden2016deep}
Daniel Holden, Jun Saito, and Taku Komura.
\newblock A deep learning framework for character motion synthesis and editing.
\newblock {\em ACM Transactions on Graphics (TOG)}, 35(4):1--11, 2016.

\bibitem{holden2015learning}
Daniel Holden, Jun Saito, Taku Komura, and Thomas Joyce.
\newblock Learning motion manifolds with convolutional autoencoders.
\newblock In {\em SIGGRAPH Asia 2015 Technical Briefs}, pages 1--4. 2015.

\bibitem{kanazawa2018end}
Angjoo Kanazawa, Michael~J Black, David~W Jacobs, and Jitendra Malik.
\newblock End-to-end recovery of human shape and pose.
\newblock In {\em Proceedings of the IEEE conference on computer vision and
  pattern recognition}, pages 7122--7131, 2018.

\bibitem{kania2021trajevae}
Kacper Kania, Marek Kowalski, and Tomasz Trzci{\'n}ski.
\newblock Trajevae--controllable human motion generation from trajectories.
\newblock {\em arXiv preprint arXiv:2104.00351}, 2021.

\bibitem{kingma2013auto}
Diederik~P Kingma and Max Welling.
\newblock Auto-encoding variational {Bayes}.
\newblock In {\em International Conference on Learning Representions}, 2014.

\bibitem{kovar2008motion}
Lucas Kovar, Michael Gleicher, and Fr{\'e}d{\'e}ric Pighin.
\newblock Motion graphs.
\newblock In {\em ACM SIGGRAPH 2008 classes}, pages 1--10. 2008.

\bibitem{lavalle1998rapidly}
Steven~M LaValle et~al.
\newblock Rapidly-exploring random trees: A new tool for path planning.
\newblock 1998.

\bibitem{lavalle2001rapidly}
Steven~M LaValle, James~J Kuffner, BR Donald, et~al.
\newblock Rapidly-exploring random trees: Progress and prospects.
\newblock {\em Algorithmic and computational robotics: new directions},
  5:293--308, 2001.

\bibitem{lee2021learning}
Kyungho Lee, Sehee Min, Sunmin Lee, and Jehee Lee.
\newblock Learning time-critical responses for interactive character control.
\newblock {\em ACM Transactions on Graphics (TOG)}, 40(4):1--11, 2021.

\bibitem{li2018convolutional}
Chen Li, Zhen Zhang, Wee Sun~Lee, and Gim Hee~Lee.
\newblock Convolutional sequence to sequence model for human dynamics.
\newblock In {\em IEEE Conference on Computer Vision and Pattern Recognition},
  pages 5226--5234, 2018.

\bibitem{li2020dynamic}
Maosen Li, Siheng Chen, Yangheng Zhao, Ya Zhang, Yanfeng Wang, and Qi Tian.
\newblock Dynamic multiscale graph neural networks for {3D} skeleton based
  human motion prediction.
\newblock In {\em IEEE Conference on Computer Vision and Pattern Recognition},
  pages 214--223, 2020.

\bibitem{li2021learn}
Ruilong Li, Shan Yang, David~A. Ross, and Angjoo Kanazawa.
\newblock Ai choreographer: Music conditioned 3d dance generation with aist++.
\newblock In {\em ICCV}, 2021.

\bibitem{ling2020character}
Hung~Yu Ling, Fabio Zinno, George Cheng, and Michiel Van De~Panne.
\newblock Character controllers using motion {VAEs}.
\newblock {\em ACM Transactions on Graphics}, 39(4):40--1, 2020.

\bibitem{liu20204d}
Miao Liu, Dexin Yang, Yan Zhang, Zhaopeng Cui, James~M Rehg, and Siyu Tang.
\newblock 4d human body capture from egocentric video via 3d scene grounding.
\newblock In {\em 2020 International Conference on 3D Vision (3DV)}. IEEE,
  2021.

\bibitem{liu2021motion}
Zhenguang Liu, Pengxiang Su, Shuang Wu, Xuanjing Shen, Haipeng Chen, Yanbin
  Hao, and Meng Wang.
\newblock Motion prediction using trajectory cues.
\newblock In {\em Proceedings of the IEEE/CVF International Conference on
  Computer Vision}, pages 13299--13308, 2021.

\bibitem{AMASS:ICCV:2019}
Naureen Mahmood, Nima Ghorbani, Nikolaus~F. Troje, Gerard Pons-Moll, and
  Michael~J. Black.
\newblock {AMASS}: Archive of motion capture as surface shapes.
\newblock In {\em International Conference on Computer Vision}, pages
  5442--5451, Oct. 2019.

\bibitem{Mandery2016b}
Christian Mandery, \"Omer Terlemez, Martin Do, Nikolaus Vahrenkamp, and Tamim
  Asfour.
\newblock Unifying representations and large-scale whole-body motion databases
  for studying human motion.
\newblock {\em IEEE Transactions on Robotics}, 32(4):796--809, 2016.

\bibitem{mao2021generating}
Wei Mao, Miaomiao Liu, and Mathieu Salzmann.
\newblock Generating smooth pose sequences for diverse human motion prediction.
\newblock In {\em Proceedings of the IEEE/CVF International Conference on
  Computer Vision}, pages 13309--13318, 2021.

\bibitem{mao2019learning}
Wei Mao, Miaomiao Liu, Mathieu Salzmann, and Hongdong Li.
\newblock Learning trajectory dependencies for human motion prediction.
\newblock In {\em International Conference on Computer Vision}, pages
  9489--9497, 2019.

\bibitem{martinez2017human}
Julieta Martinez, Michael~J Black, and Javier Romero.
\newblock On human motion prediction using recurrent neural networks.
\newblock In {\em IEEE Conference on Computer Vision and Pattern Recognition},
  pages 2891--2900, 2017.

\bibitem{merel2018neural}
Josh Merel, Leonard Hasenclever, Alexandre Galashov, Arun Ahuja, Vu Pham, Greg
  Wayne, Yee~Whye Teh, and Nicolas Heess.
\newblock Neural probabilistic motor primitives for humanoid control.
\newblock In {\em International Conference on Learning Representations}, 2019.

\bibitem{muller2007documentation}
Meinard M{\"u}ller, Tido R{\"o}der, Michael Clausen, Bernhard Eberhardt,
  Bj{\"o}rn Kr{\"u}ger, and Andreas Weber.
\newblock Documentation mocap database {HDM05}.
\newblock 2007.

\bibitem{STAR:ECCV:2020}
Ahmed A.~A. Osman, Timo Bolkart, and Michael~J. Black.
\newblock {STAR}: Sparse trained articulated human body regressor.
\newblock In {\em European Conference on Computer Vision}, volume LNCS 12355,
  pages 598--613, Aug. 2020.

\bibitem{SMPL-X:2019}
Georgios Pavlakos, Vasileios Choutas, Nima Ghorbani, Timo Bolkart, Ahmed A.~A.
  Osman, Dimitrios Tzionas, and Michael~J. Black.
\newblock Expressive body capture: {3D} hands, face, and body from a single
  image.
\newblock In {\em IEEE Conference on Computer Vision and Pattern Recognition},
  pages 10975--10985, June 2019.

\bibitem{pavllo2019modeling}
Dario Pavllo, Christoph Feichtenhofer, Michael Auli, and David Grangier.
\newblock Modeling human motion with quaternion-based neural networks.
\newblock {\em International Journal of Computer Vision}, pages 1--18, 2019.

\bibitem{pavllo2018quaternet}
Dario Pavllo, David Grangier, and Michael Auli.
\newblock Quaternet: A quaternion-based recurrent model for human motion.
\newblock In {\em British Machine Vision Conference}, 2018.

\bibitem{2018-TOG-deepMimic}
Xue~Bin Peng, Pieter Abbeel, Sergey Levine, and Michiel van~de Panne.
\newblock Deepmimic: Example-guided deep reinforcement learning of
  physics-based character skills.
\newblock {\em ACM Trans. Graph.}, 37(4):143:1--143:14, July 2018.

\bibitem{peng2018sfv}
Xue~Bin Peng, Angjoo Kanazawa, Jitendra Malik, Pieter Abbeel, and Sergey
  Levine.
\newblock Sfv: Reinforcement learning of physical skills from videos.
\newblock {\em ACM Transactions On Graphics (TOG)}, 37(6):1--14, 2018.

\bibitem{peng2021amp}
Xue~Bin Peng, Ze Ma, Pieter Abbeel, Sergey Levine, and Angjoo Kanazawa.
\newblock Amp: Adversarial motion priors for stylized physics-based character
  control.
\newblock {\em arXiv preprint arXiv:2104.02180}, 2021.

\bibitem{petrovich21actor}
Mathis Petrovich, Michael~J. Black, and G{\"u}l Varol.
\newblock Action-conditioned 3{D} human motion synthesis with transformer
  {VAE}.
\newblock In {\em International Conference on Computer Vision (ICCV)}, pages
  10985--10995, October 2021.

\bibitem{rempe2021humor}
Davis Rempe, Tolga Birdal, Aaron Hertzmann, Jimei Yang, Srinath Sridhar, and
  Leonidas~J. Guibas.
\newblock Humor: 3d human motion model for robust pose estimation.
\newblock In {\em International Conference on Computer Vision (ICCV)}, 2021.

\bibitem{MANO:SIGGRAPHASIA:2017}
Javier Romero, Dimitrios Tzionas, and Michael~J. Black.
\newblock {Embodied Hands}: Modeling and capturing hands and bodies together.
\newblock {\em ACM Transactions on Graphics, (Proc. SIGGRAPH Asia)}, 36(6),
  Nov. 2017.

\bibitem{schulman2017proximal}
John Schulman, Filip Wolski, Prafulla Dhariwal, Alec Radford, and Oleg Klimov.
\newblock Proximal policy optimization algorithms.
\newblock {\em arXiv preprint arXiv:1707.06347}, 2017.

\bibitem{sigal2010humaneva}
Leonid Sigal, Alexandru~O Balan, and Michael~J Black.
\newblock {HumanEva}: Synchronized video and motion capture dataset and
  baseline algorithm for evaluation of articulated human motion.
\newblock {\em International Journal of Computer Vision}, 87(1-2):4, 2010.

\bibitem{snook2000simplified}
Greg Snook.
\newblock Simplified 3d movement and pathfinding using navigation meshes.
\newblock {\em Game programming gems}, 1(1):288--304, 2000.

\bibitem{sohn2015learning}
Kihyuk Sohn, Honglak Lee, and Xinchen Yan.
\newblock Learning structured output representation using deep conditional
  generative models.
\newblock {\em Advances in neural information processing systems},
  28:3483--3491, 2015.

\bibitem{starke2019neural}
Sebastian Starke, He Zhang, Taku Komura, and Jun Saito.
\newblock Neural state machine for character-scene interactions.
\newblock {\em ACM Trans. Graph.}, 38(6):209--1, 2019.

\bibitem{sutton1998introduction}
Richard~S Sutton, Andrew~G Barto, et~al.
\newblock {\em Introduction to reinforcement learning}, volume 135.
\newblock MIT press Cambridge, 1998.

\bibitem{tang2018long}
Yongyi Tang, Lin Ma, Wei Liu, and Wei-Shi Zheng.
\newblock Long-term human motion prediction by modeling motion context and
  enhancing motion dynamic.
\newblock In {\em International Joint Conference on Artificial Intelligence},
  page 935–941, 2018.

\bibitem{teichner1954recent}
Warren~H Teichner.
\newblock Recent studies of simple reaction time.
\newblock {\em Psychological Bulletin}, 51(2):128, 1954.

\bibitem{trumble2017total}
Matthew Trumble, Andrew Gilbert, Charles Malleson, Adrian Hilton, and John~P
  Collomosse.
\newblock Total capture: 3d human pose estimation fusing video and inertial
  sensors.
\newblock In {\em BMVC}, volume~2, pages 1--13, 2017.

\bibitem{walker2017pose}
Jacob Walker, Kenneth Marino, Abhinav Gupta, and Martial Hebert.
\newblock The pose knows: Video forecasting by generating pose futures.
\newblock In {\em International Conference on Computer Vision}, pages
  3332--3341, 2017.

\bibitem{wang2021synthesizing}
Jiashun Wang, Huazhe Xu, Jingwei Xu, Sifei Liu, and Xiaolong Wang.
\newblock Synthesizing long-term 3d human motion and interaction in 3d scenes.
\newblock In {\em Proceedings of the IEEE/CVF Conference on Computer Vision and
  Pattern Recognition}, pages 9401--9411, 2021.

\bibitem{wang2019combining}
Zhiyong Wang, Jinxiang Chai, and Shihong Xia.
\newblock Combining recurrent neural networks and adversarial training for
  human motion synthesis and control.
\newblock {\em IEEE transactions on visualization and computer graphics},
  27(1):14--28, 2019.

\bibitem{wei2020his}
Mao Wei, Liu Miaomiao, and Salzemann Mathieu.
\newblock History repeats itself: Human motion prediction via motion attention.
\newblock In {\em European Conference on Computer Vision}, 2020.

\bibitem{xu2020ghum}
Hongyi Xu, Eduard~Gabriel Bazavan, Andrei Zanfir, William~T Freeman, Rahul
  Sukthankar, and Cristian Sminchisescu.
\newblock {GHUM} \& {GHUML}: Generative {3D} human shape and articulated pose
  models.
\newblock In {\em IEEE Conference on Computer Vision and Pattern Recognition},
  pages 6184--6193, 2020.

\bibitem{yan2018mt}
Xinchen Yan, Akash Rastogi, Ruben Villegas, Kalyan Sunkavalli, Eli Shechtman,
  Sunil Hadap, Ersin Yumer, and Honglak Lee.
\newblock {MT-VAE}: Learning motion transformations to generate multimodal
  human dynamics.
\newblock In {\em European Conference on Computer Vision}, pages 265--281,
  2018.

\bibitem{yuan2020dlow}
Ye Yuan and Kris Kitani.
\newblock {DLow}: Diversifying latent flows for diverse human motion
  prediction.
\newblock {\em European Conference on Computer Vision}, 2020.

\bibitem{yuan2021simpoe}
Ye Yuan, Shih-En Wei, Tomas Simon, Kris Kitani, and Jason Saragih.
\newblock Simpoe: Simulated character control for 3d human pose estimation.
\newblock In {\em Proceedings of the IEEE/CVF Conference on Computer Vision and
  Pattern Recognition}, pages 7159--7169, 2021.

\bibitem{zhang2021learning}
Siwei Zhang, Yan Zhang, Federica Bogo, Marc Pollefeys, and Siyu Tang.
\newblock Learning motion priors for 4d human body capture in 3d scenes.
\newblock In {\em Proceedings of the IEEE/CVF International Conference on
  Computer Vision}, pages 11343--11353, 2021.

\bibitem{zhang2020perpetual}
Yan Zhang, Michael~J Black, and Siyu Tang.
\newblock Perpetual motion: Generating unbounded human motion.
\newblock {\em arXiv preprint arXiv:2007.13886}, 2020.

\bibitem{zhang2021we}
Yan Zhang, Michael~J Black, and Siyu Tang.
\newblock We are more than our joints: Predicting how 3d bodies move.
\newblock In {\em Proceedings of the IEEE/CVF Conference on Computer Vision and
  Pattern Recognition}, pages 3372--3382, 2021.

\bibitem{zhang2020generating}
Yan Zhang, Mohamed Hassan, Heiko Neumann, Michael~J Black, and Siyu Tang.
\newblock Generating 3d people in scenes without people.
\newblock In {\em Proceedings of the IEEE/CVF Conference on Computer Vision and
  Pattern Recognition}, pages 6194--6204, 2020.

\bibitem{zhou2019continuity}
Yi Zhou, Connelly Barnes, Jingwan Lu, Jimei Yang, and Hao Li.
\newblock On the continuity of rotation representations in neural networks.
\newblock In {\em IEEE Conference on Computer Vision and Pattern Recognition},
  pages 5745--5753, 2019.

\bibitem{zhou2018autoconditioned}
Yi Zhou, Zimo Li, Shuangjiu Xiao, Chong He, Zeng Huang, and Hao Li.
\newblock Auto-conditioned recurrent networks for extended complex human motion
  synthesis.
\newblock In {\em International Conference on Learning Representions}, 2018.

\bibitem{zinno2019ml}
Fabio Zinno.
\newblock Ml tutorial day: From motion matching to motion synthesis, and all
  the hurdles in between.
\newblock {\em Proc. of GDC 2019}, 2019.

\end{thebibliography}
}

\newpage
\begingroup

\appendix
\onecolumn

\begin{center}
\Large{\bf The Wanderings of Odysseus in 3D Scenes \\ **Appendix**}
\end{center}

\setcounter{page}{1}
\setcounter{table}{0}
\setcounter{figure}{0}
\renewcommand{\thetable}{S\arabic{table}}
\renewcommand{\thefigure}{S\arabic{figure}}

\section{Additional Experiments}

\subsection{Analysis on GAMMA}
\label{sec:experiment:exp3}
\paragraph{Marker predictor analysis.}
The marker predictor performances are tested on motion primitives from {\bf ACCAD} and {\bf HumanEva}.
Following the metrics in~\cite{yuan2020dlow, zhang2021we}, the evaluation is based on prediction diversity and accuracy. 
The diversity score is the higher the better, and other scores measuring the prediction errors are the lower the better.
The results are shown in Tab.~\ref{tab:predictor_test}.
The 1-frame-based predictors have much larger diversity, but lower accuracies than the 2-frame-based predictors, because generating motion from a static pose is more uncertain. 
Moreover, the rollout training can slightly improve diversity. 
Note that our work focuses on motion generation, and hence the accuracy is less important.

\begin{table*}[h]
    \footnotesize
    \centering
    \begin{tabular}{lllllllllll}
        \toprule
        &\multicolumn{5}{c}{\textbf{ACCAD}~\cite{accad},\textit{  20 sub., 6,298 MPs}} & \multicolumn{5}{c}{\textbf{HumanEva}~\cite{sigal2010humaneva},\textit{  3 sub., 2021 MPs}}\\
        \cmidrule(lr){2-6} \cmidrule(lr){7-11}
        Method & \textit{Diversity}$\uparrow$ & \textit{ADE}$\downarrow$ & \textit{FDE}$\downarrow$ & \textit{MMADE}$\downarrow$ & \textit{MMFDE}$\downarrow$  & \textit{Diversity}$\uparrow$ & \textit{ADE}$\downarrow$ & \textit{FDE}$\downarrow$ & \textit{MMADE}$\downarrow$ & \textit{MMFDE}$\downarrow$  \\
        \midrule
        predictor-1f & 1.64 & 0.42& 0.69& 0.57& 0.81 & 2.02& 0.30& 0.48& 0.42& 0.55\\
        predictor-2f & 0.50 & 0.19& 0.34& 0.44& 0.60 & 0.53& 0.16& 0.28& 0.39& 0.48\\
        predictor-ro-1f & 1.72 & 0.44& 0.72& 0.58 & 0.83& 2.08 & 0.33& 0.51& 0.44& 0.57\\  
        predictor-ro-2f & 0.88 & 0.21& 0.36& 0.44& 0.59 & 0.95& 0.18& 0.31& 0.37& 0.46\\
        \bottomrule
    \end{tabular}
    \caption{Analysis on marker predictors. `-xf' denotes the motion seed has x frames. `-ro' denotes the model is fine-tuned with rollout. `sub' and `MPs' denote number of subjects and motion primitives, respectively. } 
    \label{tab:predictor_test}
\end{table*}

Fig.~\ref{fig:pp_analysis} illustrates the difference between the one frame-based and the two frame-based models, respectively. 
Provided a single frame, the generated motion primitive is more diverse than the generated motion based on two frames. 
Because of the motion dynamics and the body inertia, the motion becomes more deterministic.
We also tried predicting 7 frames based on 3 frames, but did not observe obvious differences with the 2-frame-based generation.

\begin{figure}[h]
    \centering
    \includegraphics[width=0.6\linewidth]{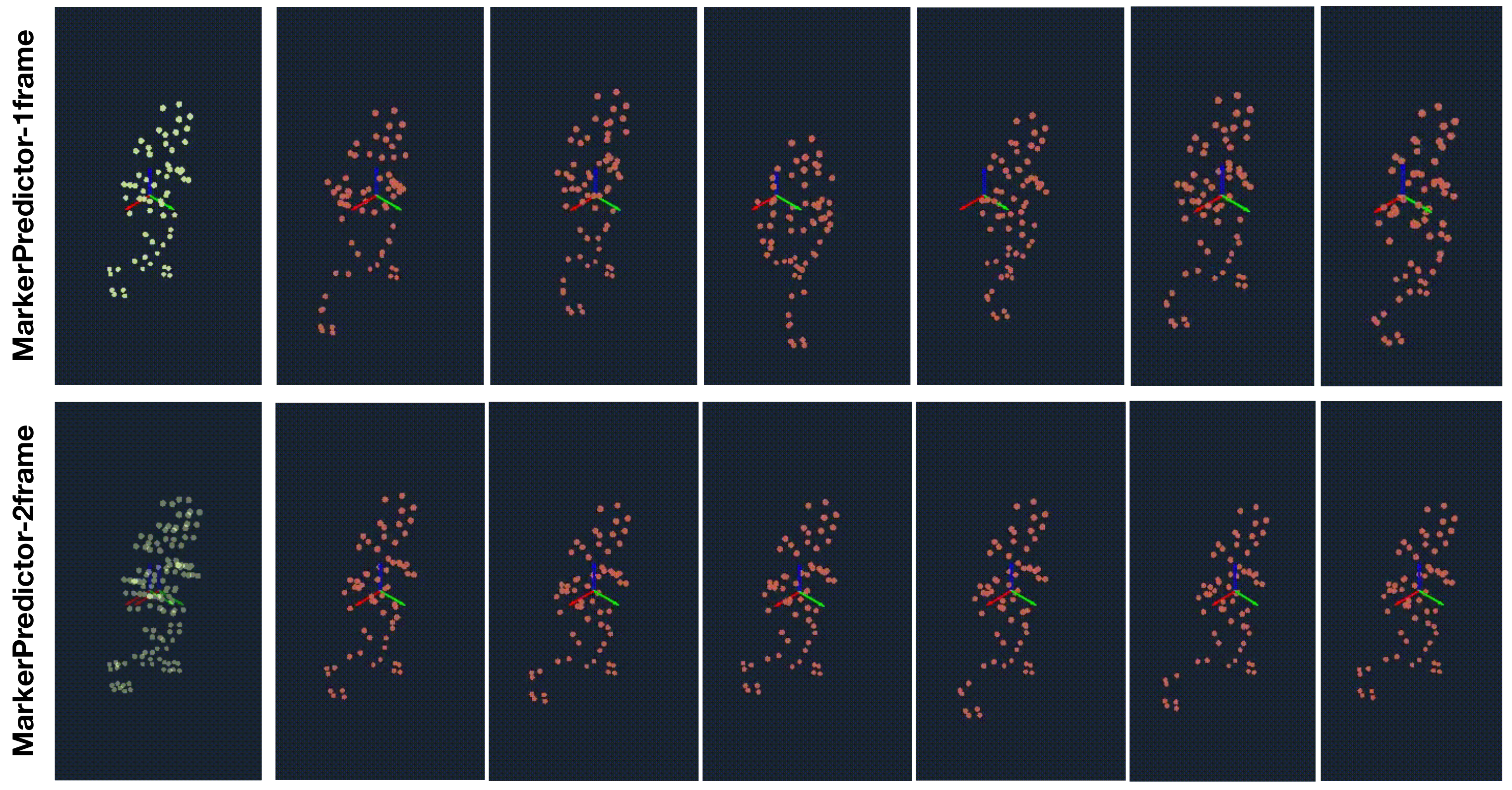}
    \caption{Visualization of generated motion primitives. In each row, the motion seed (green) is on the left, and the last frames of six generated primitives (red) are on the right.}
    \label{fig:pp_analysis}
\end{figure}

\paragraph{Body regressor analysis.}
\label{sec:regressor_test}
The performance is tested on motion primitives from {\bf HumanEva}, {\bf SSM}, {\bf ACCAD} and {\bf KIT}. 
We calculate the averaged marker distance (AMD) and the averaged mesh vertices distance (AVD) to the ground truth motion primitive for accuracy evaluation. 
The results are shown in Tab.~\ref{tab:regressor_test}. One can see that our proposed body regressors are accurate, and achieve comparable performance with Mosh and Mosh++, e.g. \cite[Sec. 4.4]{AMASS:ICCV:2019}.

\begin{table}[h]
    \centering
    \setlength\tabcolsep{1.5pt}
    \begin{tabular}{lcccc}
        \toprule
          &\multicolumn{2}{c}{male} 
          & \multicolumn{2}{c}{female}
          \\
        \cmidrule(lr){2-3} \cmidrule(lr){4-5} 
        & AMD(mm) & AVD(mm) & AMD(mm) & AVD(mm) \\
        \midrule
        {{\bf HumanEva} \textit{3 sub., 2021 MPs}} & 3.26 & 4.80 & 3.52 & 5.52\\
        {{\bf SSM} \textit{3 sub., 285 MPs}} & - & - & 6.23 & 8.24\\
        {{\bf ACCAD} \textit{20 sub., 6,298 MPs}} & 7.99 & 9.61 & 11.39 & 12.04\\
        {{\bf KIT} \textit{55 sub., 130,331 MPs}} & 3.53 & 4.88 & 3.95 & 5.64\\
        \bottomrule
    \end{tabular}
    \vspace{-0.1in}
   \caption{Evaluation of body regressors w.r.t. reconstruction errors. }
    \vspace{-1mm}
    \label{tab:regressor_test}
\end{table}

\subsection{Influence of Bodies on Motion Generation}
First, we fix the initial pose and the motion target, and interpolate the first two components from $-6$ to $6$ in the shape space to obtain 10 bodies. 
We find all bodies can reach the target, with the minimal/maximal steps being 12/53. The first row of Fig.~\ref{fig:rebuttal_shapepose} shows 4 corner cases. These similar actions indicate the limited influence of the body shape.
Second, we fix the body shape and randomly sample 10 poses from VPoser~\cite{SMPL-X:2019}. 
We find 3 bodies cannot reach the target within 60 steps, indicating the initial pose can largely influence motion generation. 
If the initial pose is not about walking, the body first adjusts itself to a `ready-to-walk' pose (via stepping back, jumping, etc.), and then walks out. The second row of Fig.~\ref{fig:rebuttal_shapepose} shows 3 examples. 

\begin{figure}[h!]
    \centering
    \includegraphics[width=\linewidth]{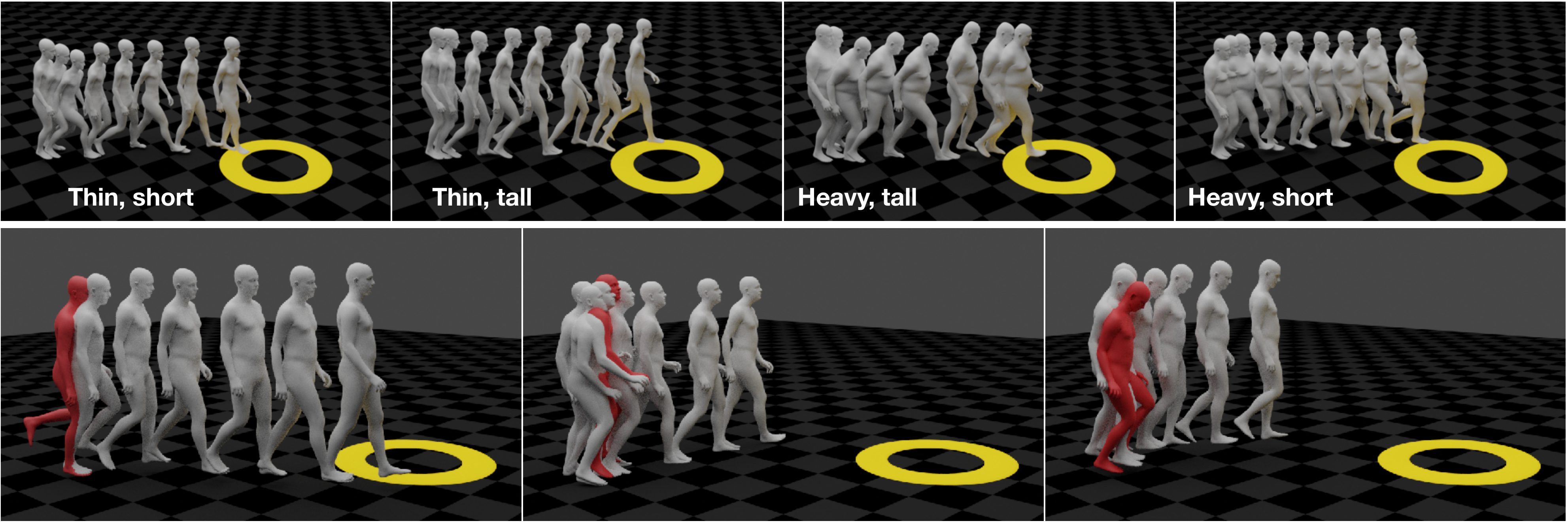}
    \caption{Influence of the body shape and the initial body pose on motion generation. See videos at \url{https://yz-cnsdqz.github.io/eigenmotion/GAMMA/}.}
    \label{fig:rebuttal_shapepose}
\end{figure}

\subsection{Comparison with MOJO~\cite{zhang2021we}}
Both our method and MOJO use the marker-based body representation and RNN-based neural networks for motion modelling. 
In contrast to long-term motion synthesis, MOJO~\cite{zhang2021we} is proposed for stochastic motion prediction and uses a different setting than our method GAMMA. 
Specifically, MOJO is exploited to predict 3-second motions with a 1-second motion seed.
Therefore, we only perform comparison w.r.t. the motion realism, e.g.~\cite[Table 4]{zhang2021we}.

Here, we randomly select 900 static bodies from {\bf ACCAD}, and generate 20 primitives ($\sim 5$ sec) for each body. 
Using the same foot skating metric with MOJO, we obtain 0.064, which is significantly better than MOJO (0.341 and 0.278) and is comparable to the ground truth (0.067).
In addition, Fig.~\ref{fig:random_gen} in our paper shows that our results have similar perceptual scores with the ground truth, whereas \cite[Table 4]{zhang2021we} shows a much larger gap between the generated motion and the ground truth. 
These results can indicate the advantage of our motion primitive representation.

\subsection{Comparison with Other Motion Representations}
Our proposed method GAMMA comprises a marker predictor to generate future markers and a body regressor to recover 3D bodies from predicted markers. 
Therefore, the motion generation is performed in an `indirect' manner. In contrast, a direct approach is to generate SMPL-X body parameters without any intermediate body representations. In this case, we can only use a single neural network to model their temporal relations.

To investigate their performance, here we use two body representations in the motion primitive setting, 
i.e. 1) the SMPL parameters ${\bm x} = ( r, \Phi,\theta)$, and 2) the HuMoR kinematic feature~\cite{rempe2021humor} ${\bm x} = ( r,\dot{r}, \Phi, \dot{\Phi},\theta, J, \dot{J})$, in which $r$, $\Phi$, $\theta$, $J$ denotes the root translation, orientation, joint rotations, and joint locations, respectively. 
All rotations are in axis-angle. 
We denote them as `SMPL params' and `HuMoR primitive', respectively.

Since we can directly obtain 3D bodies by inputting these features into SMPL-X~\cite{SMPL-X:2019}, we only use a modified version of the marker predictor (see Fig.~\ref{fig:network}) for motion modeling. Except for the input and output dimensions, other network hyper-parameters are the same for a fair comparison.
The training procedures follow Sec.~\ref{sec:model_training}, i.e., first training with individual motion primitives and then training with the rollout.
Moreover, we replace the feature reconstruction loss with a forward kinematic loss as in~\cite{rempe2021humor}, since directly reconstructing these features can hardly produce valid results.

We measure contact scores as in Fig.~\ref{fig:random_gen}, and the body deformation w.r.t. the temporal variation (i.e. std) of piece-wise distances of rigid body markers (see [Tab.3, MOJO]). 
In this case, the contact score is the higher the better, and the body deformation score is the lower the better.
The results are in Fig.~\ref{fig:rebuttal_bodyrepr}.
Compared to `HuMoR', other models on motion primitives have better contact scores. In particular, `HuMoR primitive' obviously outperforms `HuMoR'. This can indicate the advantage of the motion primitive setting for improving motion realism. 
In addition, compared to `HuMoR primitive' and `SMPL params', our methods perform better w.r.t. the body deformation. Invalid results, e.g., twisted bodies, are observed in their visualizations, but not observed in our results. This can indicate the advantage of the marker-based representation for constraining body DoFs.

In the literature, a number of various motion representations have been proposed. A thorough comparison between motion representations is interesting to explore, but is out of the scope of this paper.


\begin{figure}[h]
    \centering
    \includegraphics[width=\linewidth]{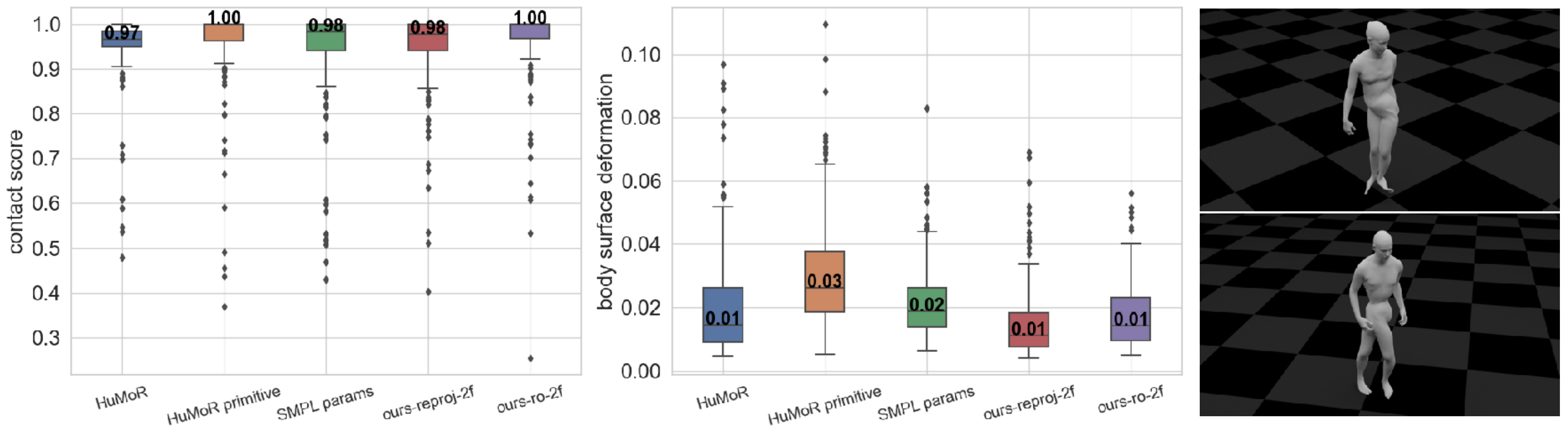}
    \caption{From left to right: contact scores, rigid body deformation measures, and typical failure cases of SMPL parameters (top) and HuMoR feature (bottom). Except `HuMoR primitive' and `SMPL params', other results are from Fig.~\ref{fig:random_gen}.}
    \label{fig:rebuttal_bodyrepr}
\end{figure}

\subsection{Runtime Test}
We perform a runtime test with a single Nvidia Quadro 6000 GPU. 
We generate motions in parallel for a batch of human bodies, and the results are shown in Tab.~\ref{tab:runtime}. 
This shows that our algorithm can produce motion efficiently. Note that each motion primitive spans 0.25 seconds, and therefore our method can simultaneously drive 256 SMPL-X bodies to move in real-time.
\begin{table}[h]
    \centering
    \begin{tabular}{lccccc}
    \toprule
       num. of bodies in a batch & 1024 & 512 & 256 & 128 & 64 \\
        \midrule
        runtime (sec. per primitive) & 0.85 & 0.5 & 0.25 & 0.15 & 0.1\\
    \bottomrule
    \end{tabular}
    \caption{Runtime test on a single GPU.}
    \label{tab:runtime}
\end{table}

\section{More demonstrations on Methods}

\subsection{End-to-end Training of Generative Motion Primitives}
Since all modules are differentiable, end-to-end training of GAMMA can be performed by minimizing
\begin{equation}
    \mathcal{L} = \mathbb{E}_{{\bm \Theta}}[|M\circ \mathcal{M}({\bm \Theta}, {\bm \beta})-{\bm Y}|] + \mathbb{E}_{{\bm \Theta}}[|M\circ\Delta\mathcal{M}({\bm \Theta}, {\bm \beta})-\Delta{\bm Y}|] + \Psi\left(\text{KL-div}( q({\bm Z} | {{\bm X},{\bm Y}}) || \mathcal{N}(0, {\bm I}))) \right) + \alpha |{\bm \theta^h}|^2,
    \label{eq:trainloss_end_2_end}
\end{equation}
in which the reconstruction terms are directly applied to the regressed body meshes. 
Such end-to-end training is much slower than training the marker predictor and the body regressor separately, particularly with the rollout. 

In principle, this end-to-end training can make the motion generation more stable than training modules separately. 
For example, the body regressor can handle predicted markers as input, considering the prediction error accumulation.
However, we do not observe benefits of such end-to-end training in our experiments (see Tab.~\ref{fig:random_gen} in Sec.~\ref{sec:experiment}).
The error accumulation problem cannot be solved thoroughly. Out-of-distribution cases can happen, causing the regressor fails and the predictor crashes.
We think MOJO's reprojection approach is more robust, due to the pose regularizer in the optimization loss. However, it runs at 2sec/frame, which is too slow for long-term motion generation. 
We propose to design a real-time and robust body optimizer in the future.

\subsection{More Demonstrations about Policy Training}
Besides the state, the action, and the reward, which are demonstrated in Sec.~\ref{sec:motion_control}, we need another two variables for the PPO method.

\paragraph{Expected return.}
In our study, the expected return at time $t$, i.e., $R_t$, is based on the reward and a discount factor $0<\gamma<1$, which is set to 0.99 in our trials. 
The expected return is calculated as
\begin{equation}
    R_t = \sum_{\tau=t}^{\infty} \gamma^{\tau-t}r_{\tau},
\end{equation}
in which the reward is defined in Eq.~\eqref{eq:reward}. 

\paragraph{Advantage function.}
The advantage function is defined as the difference between the expected return and the value function produced by the critic network. It is given by
\begin{equation}
    A({\bm s}_t, {\bm z}_t) = A_t = R_t - V({\bm s}_t).
\end{equation}
In our trials, the advantage function $A_t$ is normalized for all body clones in one simulation, which can empirically stabilize the training process.

\paragraph{The actor-critic network}
Compared to the actor (see Fig.~\ref{fig:network}), the critic network is only used during training for calculating the advantage function. It has a similar architecture to the actor and shares the GRU with it. 
The full actor-critic network is shown in Fig.~\ref{fig:critic}.

\begin{figure}[h]
    \centering
    \includegraphics[width=0.5\linewidth]{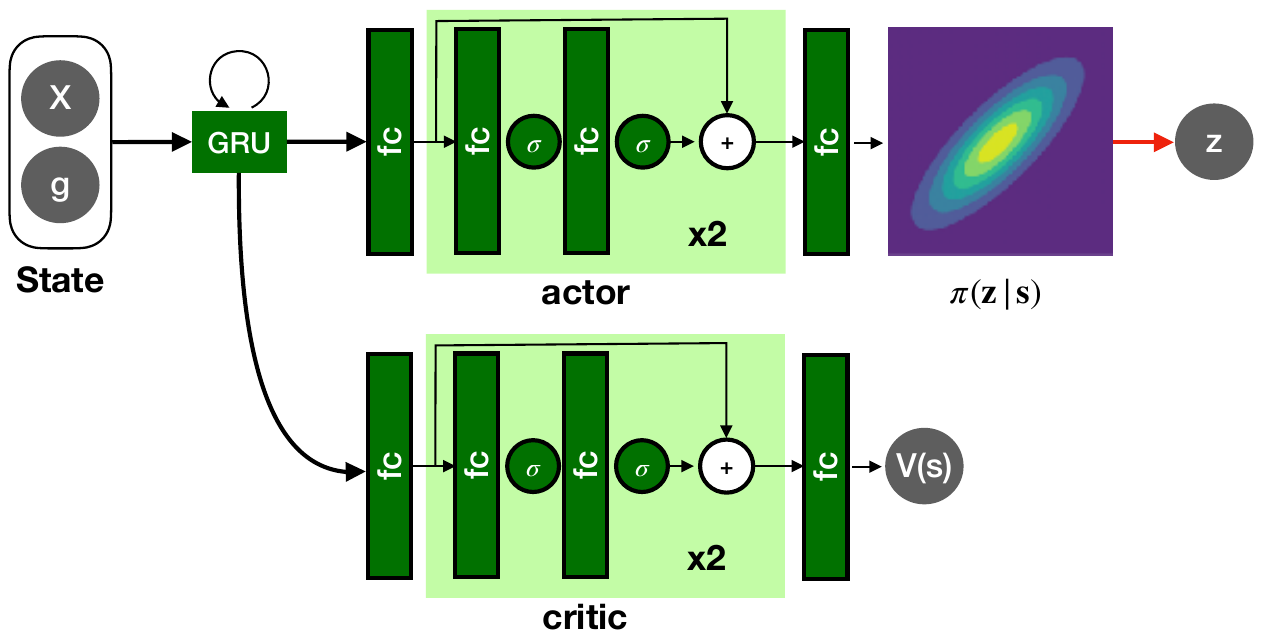}
    \caption{Architecture of the actor-critic network. Compared to the actor, the critic shares the GRU with it, and produces a scalar as the state value function.}
    \label{fig:critic}
\end{figure}

\paragraph{PPO training.}
With the advantage function, the PPO term $\mathcal{L}_{PPO}$ in Eq.\eqref{eq:policy_loss} is given by
\begin{equation}
    \mathcal{L}_{PPO} = -\min{ \left( \frac{\pi({\bm z}_t | {\bm s}_t)}{\pi^k({\bm z}_t | {\bm s}_t)} A({\bm z}_t, {\bm s}_t), \text{clip}\left(\frac{\pi({\bm z}_t | {\bm s}_t)}{\pi^k({\bm z}_t | {\bm s}_t)}, 1-\epsilon, 1+\epsilon \right) A({\bm z}_t, {\bm s}_t)  \right) },
\end{equation}
in which $\pi^k$ is a pre-calculated value in the k-th simulation, $\epsilon$ is set to 0.2 in our trials. In addition, we stop updating the policy term when the KL-divergence between $\pi$ and $\pi^k$, i.e. $\text{KLD}(\pi^k || \pi)$ is larger than 0.02. This PPO term is trained jointly with other terms.

\end{document}